\pdfoutput=1
\documentclass[11pt]{article}

\usepackage[final]{acl}

\usepackage{times}
\usepackage{latexsym}

\usepackage[T1]{fontenc}
\usepackage[utf8]{inputenc}

\usepackage{microtype}

\usepackage{inconsolata}

\usepackage{xurl}
\usepackage{booktabs}
\usepackage{adjustbox}
\usepackage{makecell}
\usepackage{multirow}
\usepackage{multicol}
\usepackage{xspace}
\usepackage[most]{tcolorbox}
\usepackage[framemethod=tikz]{mdframed}
\newcommand{\eg}{e.g.,\xspace}
\newcommand{\ie}{i.e.,\xspace}
\usepackage{newfloat}
\usepackage{listings}

\definecolor{qualcolor}{RGB}{128,64,0}
\gdef\Sepline{%
  \par\noindent\makebox[\linewidth][l]{%
  \hspace*{-\mdflength{innerleftmargin}}%
   \tikz\draw[thick,dashed,gray!60] (0,0) --%
        (\textwidth+\the\mdflength{innerleftmargin}+\the\mdflength{innerrightmargin},0);
  }\par\nobreak}

\newcommand{\bluecell}{\cellcolor{blue!10}}
\newcommand{\graycell}{\cellcolor{gray!20}}

\definecolor{blendedCC}{RGB}{159, 226, 191}
\definecolor{empathyCC}{RGB}{222, 49, 99}
\definecolor{dailyCC}{RGB}{64, 224, 208}
\definecolor{knowledgeCC}{RGB}{255, 127, 80}
\definecolor{personaCC}{RGB}{100, 149, 237}

\definecolor{codegreen}{rgb}{0,0.6,0}
\definecolor{codegray}{rgb}{0.5,0.5,0.5}
\definecolor{codepurple}{rgb}{0.58,0,0.82}
\definecolor{backcolour}{rgb}{0.95,0.95,0.92}

\lstdefinestyle{mystyle}{
    backgroundcolor=\color{backcolour},   
    commentstyle=\color{codegreen},
    keywordstyle=\color{magenta},
    numberstyle=\tiny\color{codegray},
    stringstyle=\color{codepurple},
    basicstyle=\ttfamily\footnotesize,
    breakatwhitespace=false,         
    breaklines=true,                 
    captionpos=b,                    
    keepspaces=true,                 
    numbers=left,                    
    numbersep=5pt,                  
    showspaces=false,                
    showstringspaces=false,
    showtabs=false,                  
    tabsize=2
}

\title{DialogCC: An Automated Pipeline for Creating High-Quality Multi-Modal Dialogue Dataset}

\author{Young-Jun Lee\textsuperscript{\rm 1} \hspace{0.3cm}
        Byungsoo Ko \textsuperscript{\rm 2} \hspace{0.3cm}
        Han-Gyu Kim \textsuperscript{\rm 3} \hspace{0.3cm}
        Jonghwan Hyeon \textsuperscript{\rm 1} \hspace{0.3cm}
        Ho-Jin Choi \textsuperscript{\rm 1}\\
    \textsuperscript{\rm 1} School of Computing, KAIST \hspace{0.3cm}
    \textsuperscript{\rm 2} NAVER Vision \hspace{0.3cm}
    \textsuperscript{\rm 3} NAVER Cloud Multimodal AI \\
    \texttt{\{yj2961, jonghwanhyeon, hojinc\}@kaist.ac.kr} \\ \texttt{kobiso62@gmail.com} \hspace{0.3cm} \texttt{hangyu.kim@navercorp.com}
}

\begin{document}
\maketitle

\begin{abstract}
As sharing images in an instant message is a crucial factor, there has been active research on learning an image-text multi-modal dialogue models.
However, training a well-generalized multi-modal dialogue model remains challenging due to the low quality and limited diversity of images per dialogue in existing multi-modal dialogue datasets.
In this paper, we propose an automated pipeline to construct a multi-modal dialogue dataset, ensuring both dialogue quality and image diversity without requiring minimum human effort. 
In our pipeline, to guarantee the coherence between images and dialogue, we prompt GPT-4 to infer potential image-sharing moments - specifically, the utterance, speaker, rationale, and image description. 
Furthermore, we leverage CLIP similarity to maintain consistency between aligned multiple images to the utterance.
Through this pipeline, we introduce DialogCC, a high-quality and diverse multi-modal dialogue dataset that surpasses existing datasets in terms of quality and diversity in human evaluation.
Our comprehensive experiments highlight that when multi-modal dialogue models are trained using our dataset, their generalization performance on unseen dialogue datasets is significantly enhanced. We make our source code and dataset publicly available~\footnote{\url{https://dialogcc.github.io/}}.
\end{abstract}

\section{Introduction} \label{sec:intro}

People share various images with each other when communicating via instant messaging tools. Such behavior increases social bonding (rapport) as well as engagement. The ability to share images is also necessary for a dialogue model for better bonding conversations. 
In the visual dialogue domain, the majority of previous works have focused on image-grounded dialogues, where two persons talk about given images~\cite{antol2015vqa,das2017visual,mostafazadeh2017image,shuster2018image,pasunuru2018game,kottur2019clevr,meng2020openvidial,zheng2021mmchat, shuster2020multi}. 
In practical situations, humans actively share images during conversations rather than merely talking about a given image, which is called \textit{image-sharing} behavior~\cite{lobinger2016photographs}. 
Recent studies for the image-sharing have proposed multi-modal dialogue datasets, which are constructed through the crowd-sourcing (PhotoChat~\cite{zang2021photochat}), image-text similarity with human efforts (MMDD~\cite{lee2021constructing}), or social media platform (MMDialog~\cite{feng2022mmdialog}).

\begin{figure}[t!]
    \centering
    \includegraphics[width=0.8\columnwidth]{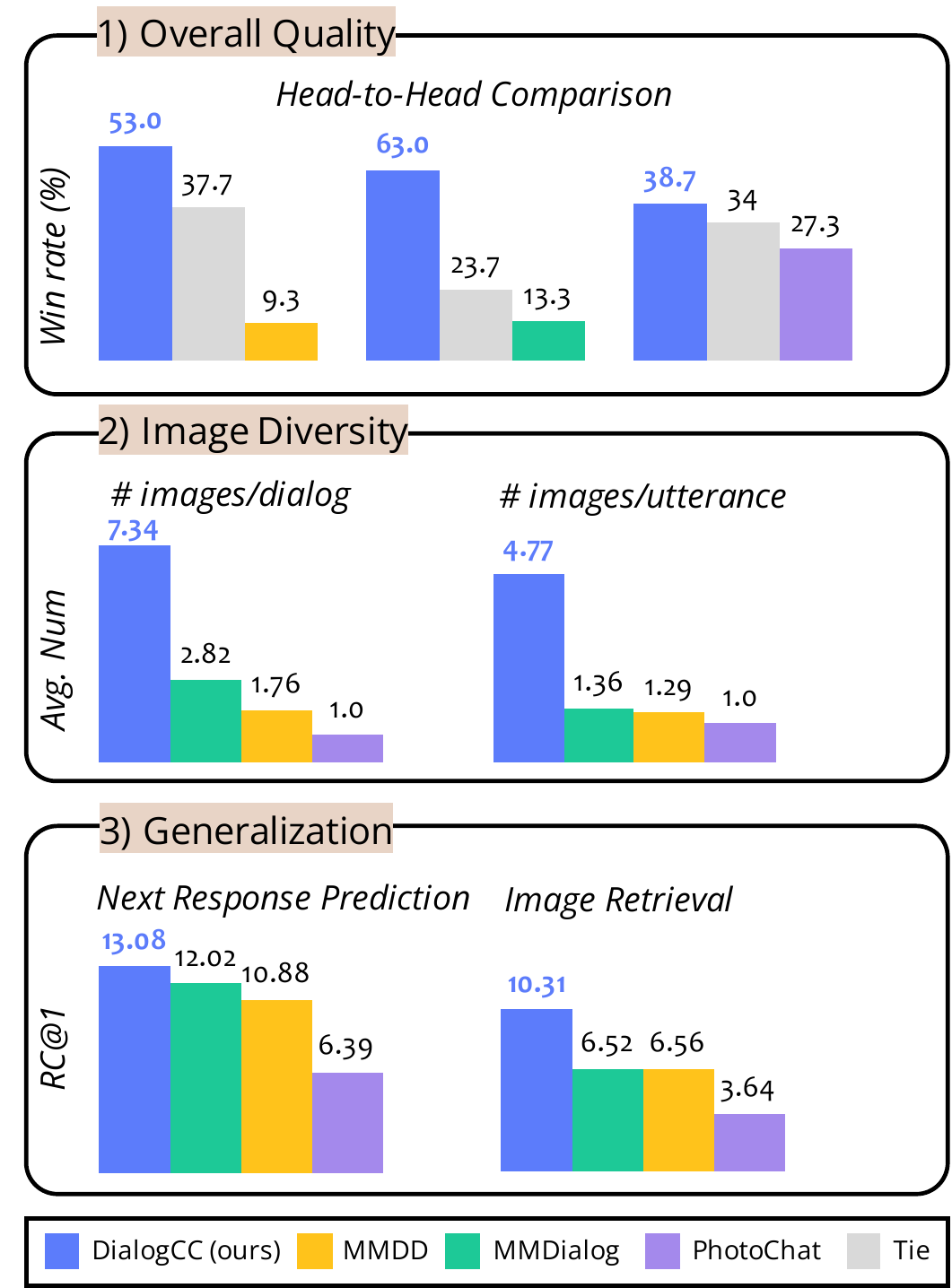}
    \caption{Comparing DialogCC (ours) to three existing multi-modal dialogue dataset in terms of a quality, diversity, and generalization. RC@1 denotes the averaged contributed R@1 performance.}
    \label{main_fig:teaser}
    \vspace{-0.5em}
\end{figure}

\begin{figure*}[!t]
    \centering
    \includegraphics[width=\textwidth]{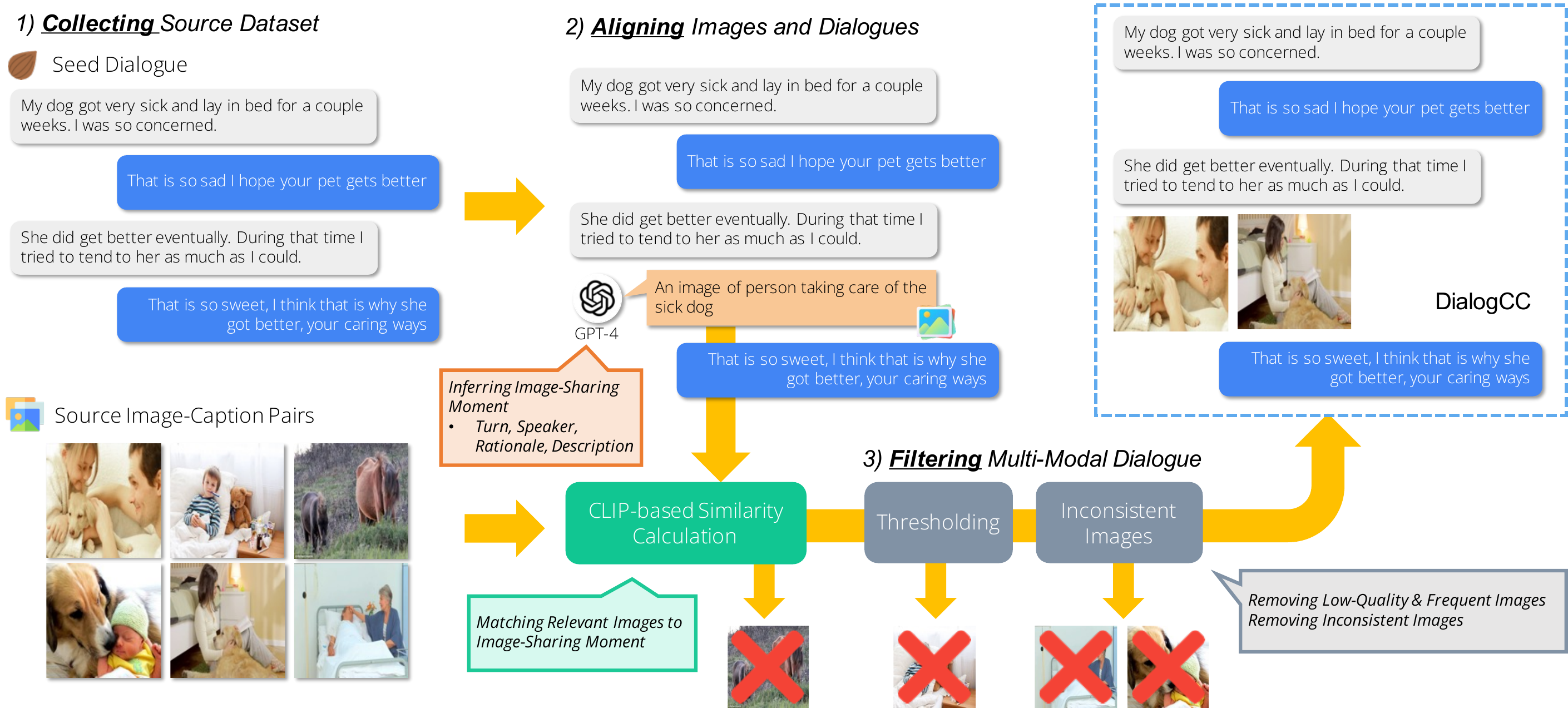}
    \caption{An overview of our proposed automatic pipeline for creating a high-quality and diverse multi-modal dialogue dataset.}
    \label{main_fig:pipeline}
    \vspace{-0.5em}
\end{figure*}

However, existing multi-modal dialogue datasets have three significant limitations; 
\textbf{(1) Quality.} Recent studies have shown that a high-quality dataset enhances both the efficacy and the quality of the model training~\cite{abbas2023semdedup,zhou2023lima}.
Nevertheless, as shown in Figure~\ref{main_fig:teaser}, existing datasets contain low-quality multi-modal dialogues (\ie appearance of images in unnatural moments, inconsistency between the image and the context of the conversation) that hinder the training process of true multi-modal social dialogue agents.
\textbf{(2) Diversity.} Given the same dialogue and context, people can share different types of images. For example, for an utterance of ``\textit{I love a dog},'' one can share an image of a chihuahua, and the other can share an image of a poodle. Nonetheless, as shown in Figure~\ref{main_fig:teaser} (\textit{\# images  / dialog} and \textit{\# images / utterance}), existing datasets consist of less than the average 2.8 images per dialogue and the average 1.4 images per utterance.
\textbf{(3) Generalization.} A model trained with conventional datasets can be overfitted by memorizing low-quality and limited pairs of images and dialogues, which can hinder its ability to handle unseen dialogue scenarios effectively by its lack of generalization. As shown in Figure~\ref{main_fig:teaser}, models trained on existing datasets show low performance on unseen dialogue datasets on both retrieval tasks. However, the model trained on our dataset achieves comparable performance, which benefited from the high quality and diversity.

This work aims to create a high-quality and diverse multi-modal dialogue dataset to train a well-generalized multi-modal dialogue model for open-domain conversation. 
To this end, we propose a fully automatic framework for creating a multi-modal dialogue dataset that involves three main steps: \textit{collecting}, \textit{aligning}, and \textit{filtering}, as shown in Figure~\ref{main_fig:pipeline}.
After collecting source datasets, to ensure image-dialogue coherence, we ask GPT-4~\cite{openai2023gpt} to infer all possible image-sharing moments via zero-shot prompting and leverage the CLIP~\cite{radford2021learning} to increase the aligned image relevancy in the \textit{aligning} step.
In the \textit{filtering} step, we eliminate inappropriate images based on CLIP similarity for image-image consistency.
We propose a high-quality and diverse multi-modal dialogue dataset, DialogCC, constructed by our proposed pipeline without minimum human efforts, unlike the previous datasets.
As illustrated in Figure~\ref{main_fig:teaser}, DialogCC achieves better statistics compared to the existing datasets in terms of quality, diversity, and generalization, indicating the effectiveness of our proposed pipeline. 
In addition, extensive experiments demonstrate that DialogCC can boost the generalization performance of trained models on unseen dialogue scenarios.

In summary, our main contributions are as follows:
1) We propose a fully automatic pipeline to create a multi-modal dialogue dataset that can achieve quality and diversity without human intervention.
2) We propose a high-quality and diverse multi-modal dialogue dataset named DialogCC, which contains various images per dialogue and utterance, respectively.
3) Extensive experiments demonstrate the effectiveness of our dataset, which enhances the generalization performance.

\section{Related Work} \label{sec:related_work}

\paragraph{Multi-Modal Dialogue Dataset.} In the visual dialogue domain, most previous studies are divided into two categories depending on whether the image is \textit{grounded} or \textit{sharing} in the dialogue. 
The image-grounded dialogue task aims to answer questions~\cite{antol2015vqa,das2017visual,seo2017visual,kottur2019clevr} or generate natural conversations~\cite{mostafazadeh2017image,shuster2018image,meng2020openvidial,wang2021openvidial,zheng2021mmchat} about given images. 
These datasets require machines to perceive and understand the given images, but we sometimes share images relevant to dialogue contexts in daily conversations. 
Hence, it is difficult to train dialogue agents to retrieve an appropriate image based on dialogue contexts in image-grounded dialogue task.

\paragraph{Image-Sharing Dialogue Dataset.} 
Recently the image-sharing dialogue task has been proposed to overcome such limitation, which predicts images semantically relevant to given dialogue contexts. 
Since there were no existing datasets for image-sharing task, previous studies have focused on construction of the dataset. 
One of the existing datasets, named PhotoChat~\cite{zang2021photochat}, is manually constructed through a crowd-sourcing platform with Open Image Dataset V4~\cite{kuznetsova2020open} as source images. 
This dataset can provide a high-quality dialogue dataset, but the manual construction is time-consuming and expensive.
Another line of work~\cite{lee2021constructing} creates a 45k multi-modal dialogue dataset by replacing an utterance with relevant images using image-text similarity, based on a threshold ensuring dialogue coherence as determined by human evaluation.
Still, we need a human-in-the-loop process and the similarity of image and utterance result is not reliable in terms of the nature of dialogue context, such as coreference resolution.
MMDialog dataset is a web-scale multi-modal dialogue dataset curated from a social media platform, but it lacks the natural conversational flow due to the nature of non-consecutive turn of social media interactions, resulting in highly low quality, which is also reported in the previous work~\cite{han2023champagne}.
All datasets cannot maintain both quality and diversity simultaneously, as demonstrated in Figure~\ref{main_fig:teaser}.
Therefore, we construct a high-quality multi-modal dialogue dataset containing various images through the proposed automatic pipeline.

\paragraph{Multi-Modal Dialogue Model.} 
The multi-modal dialogue model is mainly categorized into retrieval and generative models. The retrieval model is to retrieve proper texts or images from the candidates given the dialogue contexts. The generative model is to generate responses given the dialogue contexts. For the retrieval model, most existing studies have adopted the dual encoder architecture consisting of a text encoder and image encoder~\cite{shuster2018image,lee2021constructing,zang2021photochat}. For the generative model, many works are based on the encoder-decoder architecture~\cite{shuster2020multi,wang2021modeling,sun2021multimodal,lu2022towards}. Focusing on the \textit{image-sharing} behavior, we train a cross-modal retrieval model on our dataset, highlighting potential future applications.

\section{DialogCC} \label{sec:method}

In this section, we propose DialogCC, a high-quality and diverse multi-modal social dialogue dataset. In order to construct DialogCC, we introduce an automatic pipeline, which consists of three steps: (1) \textit{collecting}, (2) \textit{aligning}, and (3) \textit{filtering}. 
Besides, we conduct a comprehensive analysis of our dataset with respect to quality and diversity by comparing three existing datasets, MMDD~\cite{lee2021constructing}, PhotoChat~\cite{zang2021photochat}, and MMDialog~\cite{feng2022mmdialog}.
The overall pipeline is illustrated in Figure~\ref{main_fig:pipeline}. In the following part of this section, we provide details about our proposed pipeline.

\subsection{Collecting Source Dataset}

\paragraph{Source Dialogue.} As a source data, we collect five multi-turn text-only social dialogue datasets, which are publicly available online. Five dialogue datasets are Persona-Chat~\cite{zhang2018personalizing}, EmpatheticDialogues~\cite{rashkin2018towards}, Wizard-of-Wikipedia~\cite{dinan2018wizard}, DailyDialog~\cite{li2017dailydialog}, and BlendedSkillTalk~\cite{smith2020can}. 
They are manually constructed via a crowd-sourcing platform, and each dataset is specialized in specific conversational skills. 
Persona-Chat dataset contains the ability to get to know each other based on given personal information. 
EmpatheticDialogues dataset contains the ability to understand and interpret interlocutors' emotional situations and express emotional reactions adequately. 
Wizard-of-Wikipedia contains the ability to generate specific responses using knowledge or topic. 
DailyDialog contains daily life conversations with aspects, such as emotion, topic, and dialog acts. 
Lastly, in the BlendedSkillTalk, multiple skills (i.e., persona, empathy, and knowledge) are integrated into one conversation, as humans do.
We incorporate five dialogue datasets into one large dialogue dataset.

\paragraph{Source Image-Caption Pairs.} We choose Conceptual Captions 3M~\cite{sharma2018conceptual} (CC3M), which is widely used in multi-modal modeling~\cite{lu2019vilbert,su2019vl} and creating multi-modality dataset~\cite{nagrani2022learning}.
We collect 2,796,458 image-caption pairs for the training and validation set. 
Then, we discard low-quality image-caption pairs based on our filtering criteria.
First, we remove image-caption pairs with image-caption cosine similarity lower than the threshold of 0.2439 by leveraging CLIP ViT-L/14 model. 
Second, we remove watermark images using watermark detector~\footnote{\url{https://github.com/LAION-AI/LAION-5B-WatermarkDetection}}.
Lastly, we remove image-caption pairs that contain copyright-related phrases (e.g., ``royalty free'') in captions.
After the filtering, 692,292 image-caption pairs are obtained, which are divided into the training / validation / test set with a ratio of 5:1:1, resulting in 494K / 98K / 98K of unique images.
Note that our pipeline can work with any image-caption datasets, such as Conceptual Captions 12M~\cite{changpinyo2021conceptual} and RedCaps~\cite{desai2021redcaps}.

\subsection{Aligning Images and Dialogues}

After collecting a set of images and dialogues, we now describe how we create a high-quality and diverse multi-modal dialogue dataset starting from a seed text-only dialogue.

\paragraph{Inferring Image-Sharing Moments.}

While the image-sharing behavior naturally happens in existing human-authored datasets, we first should find potential image-sharing moments in the seed dialogue (Figure~\ref{main_fig:pipeline}).
However, it is challenging to determine the possible image-sharing moments in the given dialogue.
Previously, the MMDD dataset is constructed by substituting utterances and images based on cosine similarities from an image-text matching model. This method can not guarantee the quality of the image-dialogue coherency (Figure~\ref{main_fig:head_to_head}), due to the nature of multi-turn conversation.
Rather than directly measuring the similarity between utterances and images, we leverage GPT-4~\cite{openai2023gpt}~\footnote{In this work, we use GPT-4 due to the high-quality, but our pipeline could work with any LLMs, such as LLaMa-2-Chat~\cite{touvron2023llama} (See Appendix~\ref{supp_sec:open-source-llm}). We use \texttt{gpt-4-0314} version, not using the recent version \texttt{gpt-4-0613} because of the lower performance reported in~\cite{chen2023chatgpt}.} in a zero-shot setting, inspired by its recent performance in the social dialogue domain~\cite{kim2022soda,lee2022does}.
Specifically, GPT-4 infers the appropriate turn and speaker to share the image. It also provides a contextual image description and explains why the image share is appropriate. The distribution of these rationales is in Table~\ref{supp_tab:rationale_distribution}.

We use the carefully designed prompt template (detailed in the Appendix~\ref{supp_sec:prompt_template}).
To make the dialogue inputs in the prompt more natural and soundness, we use Top-10K common names of US SSN applicants from 1990 to 2021~\footnote{\url{https://catalog.data.gov/dataset/baby-names-from-social-security-card-applications-national-data}} for the speaker in a given dialogue, followed by a previous work~\cite{kim2022soda}.
However, if the original dataset contains real speaker names, this could confuse the model. To avoid this, a named entity recognizer checks for person-related entities. If none is found, we select two names from the Top-10K list. If entities are detected, we ask GPT-4 to discern the actual speaker names.
We then exclude non-human speakers, such as ``hotel'', ``corporation''.
Finally, after constructing natural dialogue, we ask the model to infer potential image-sharing moments, specifying the image-sharing utterance, speaker, rationale, and image description, in a given dialogue, with a structured format of ``\texttt{<utterance> | <speaker> | <rationale> | <image description>}''.
We parse each information in the structured format using the regex pattern (in Appendix~\ref{supp_sec:prompt_template}).

\begin{figure}
    \centering
    \includegraphics[width=\columnwidth]{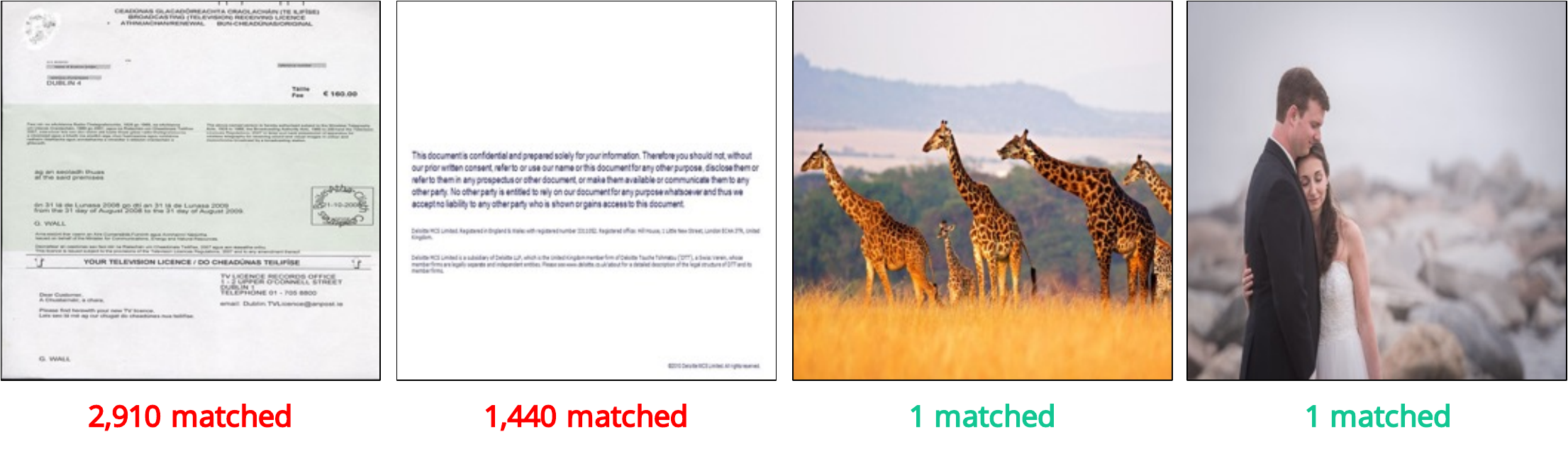}
    \caption{We show the examples of frequently matched images. The number under each image indicates the count of how many utterances are matched.}
    \label{main_fig:frequent_result}
    \vspace{-1em}
\end{figure}

\paragraph{CLIP-based Similarity Calculation.} 
In order to find images semantically relevant to a given dialogue context, we should get meaningful textual and visual features through a multi-modal feature extractor $f(\cdot)$. The previous work~\cite{lee2021constructing} used a pre-trained Visual Semantic Reasoning Network~\cite{li2019visual} as $f(\cdot)$. 
In this work, we leverage CLIP~\cite{radford2021learning} model as $f(\cdot)$, which is widely used in previous studies~\cite{bose2022movieclip,frans2021clipdraw,hessel2021clipscore,cho2022fine,zhu2023multimodal} because of a well-generalized open-domain model. 
We first extract LLM-generated description feature vector ($v_d=f(d)$), caption feature vector ($v_c=f(c)$), and image feature vector ($v_i=f(i)$). 
We then calculate the \textit{description-image} similarity by computing the cosine similarity of $v_d$ and $v_i$. 
Besides, to enhance the quality of utterance-image matching by additionally adopting the information provided by image captions, we also calculate the \textit{description-caption} similarity. 

However, there is one problem that we have to consider about how to combine these two similarity types. As reported in~\cite{liang2022mind, so2022multi}, there is a phenomenon called \textit{modality gap} in multi-modal modeling, where two different modalities (\ie image and text) are separately distributed in shared embedding space. Such phenomenon causes scale differences between description-image and description-caption similarities, so combining them directly would be biased to the larger scaled similarity. To alleviate this problem, the z-score normalization is conducted on both types of similarities, where the mean and standard deviation values for each similarity type are calculated using a training set. The normalized similarities are linearly combined as follows:
\begin{equation}
    %\mu_{\textbf{U-C}} = \sum_{u} \sum_{c} \frac{sim(v_d, v_c)}{M \times N} \\
    \mathcal{S} = \alpha f_Z\left(s_c(v_d, v_i)\right) + (1 - \alpha) f_Z\left(s_c(v_d, v_c)\right), 
    %\mathcal{S} = \alpha \text{sim}(v_u, v_i) + (1-\alpha)\text{sim}(v_u, v_c),
\end{equation}
where $s_c(x,y)$ denotes the cosine similarity and $f_Z$ represents z-score normalization.
In this paper, we set $\alpha$ as 0.5 to reflect two similarities equally. During the utterance-image matching process, the similarity matrix $\mathcal{S}$ of the size of $N \times M$ is computed, where N and M are the number of utterances and images, respectively. 
We then select the top-100 samples based on the similarity scores.

\begin{figure}
    \centering
    \includegraphics[width=\columnwidth]{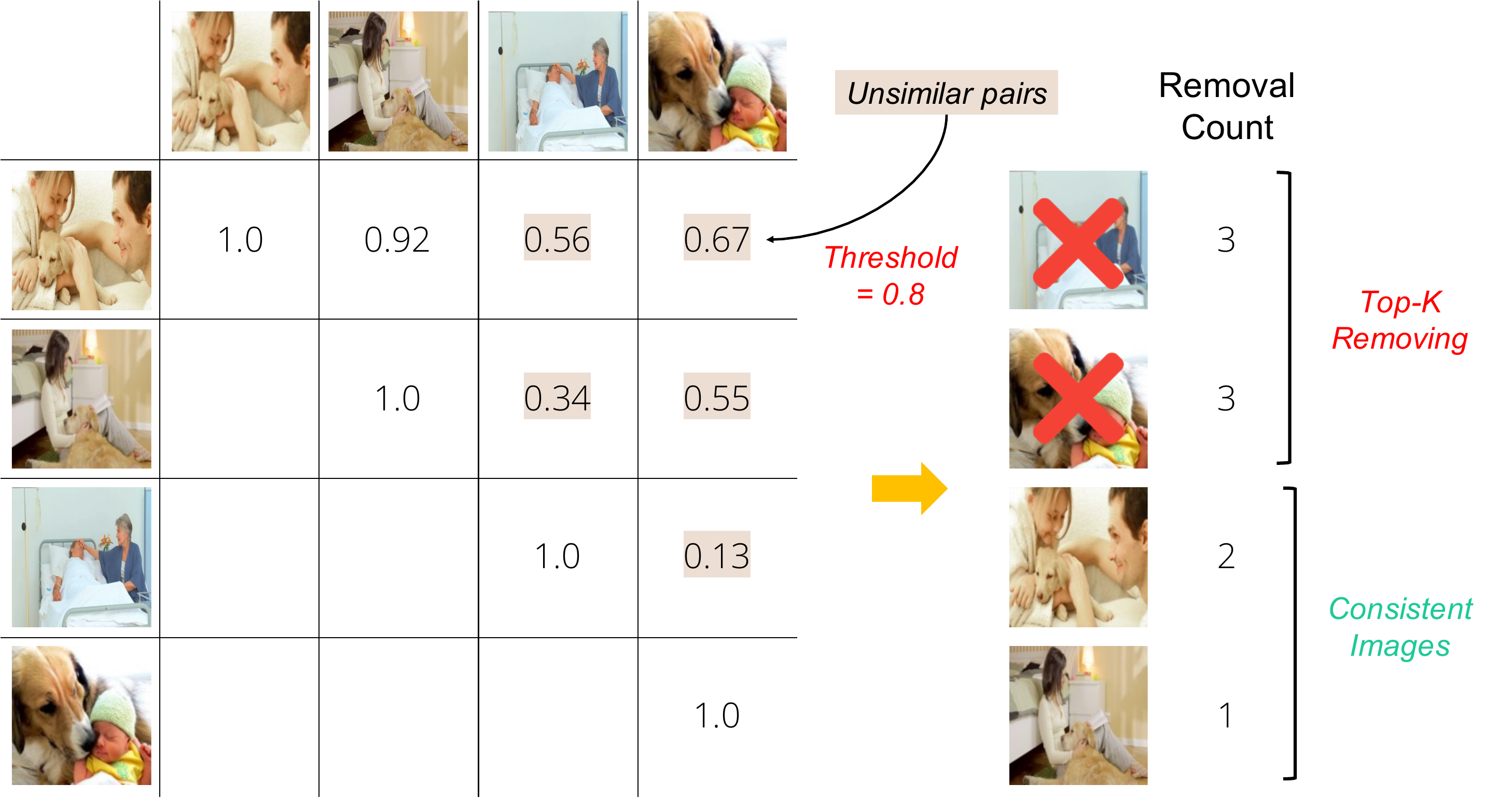}
    \caption{We illustrate the inconsistent images filtering process.}
    \label{main_fig:pairwise}
    \vspace{-1em}
\end{figure}

\subsection{Filtering Multi-Modal Dialogue}

\paragraph{Thresholding-based Filtering.} 
We have found out that there still exist unsuitable cases among the matched images found by CLIP-based similarity. 
To improve the quality of our dataset, we remove unsuitable images matched to utterances based on our criteria.
Initially, we discarded images with cosine similarity scores below 2.702, retaining only 54.05\% of the images.
Moreover, we observe that certain images are frequently matched with many utterances. As shown in Figure~\ref{main_fig:frequent_result}, the frequently matched images mostly contain textual information (\eg document) rather than object-centric or event-centric semantics (\eg ``giraffe'' or ``loving''). 
These frequent matches can lead to model overfitting, which is harmful to the generalization performance. To address this, we eliminate images that are matched more than 100 times.

\paragraph{Inconsistent Images Filtering.}
After we obtain multiple aligned images for each utterance, we should remove inconsistent images among multiple aligned images to ensure semantic similarity while maintaining diversity between multiple images. We illustrate the filtering process to help the understanding in Figure~\ref{main_fig:pairwise}.
First, we calculate a cosine similarity between multiple aligned images in a pairwise manner by leveraging the CLIP ViT-L/14 model.
Next, we regard the image pair whose similarity score is lower than the threshold $\tau$ as unsimilar pair candidate. We set $\tau$ as 0.8. Then, we increase the removal candidate count of the image in this pair by 1.
Finally, we sort by this count in descending order, and discard images in the Top-\textit{K}\% to have a high likelihood of being inconsistent with multiple images.

\begin{figure}[t]
\centering
\includegraphics[width=\columnwidth]{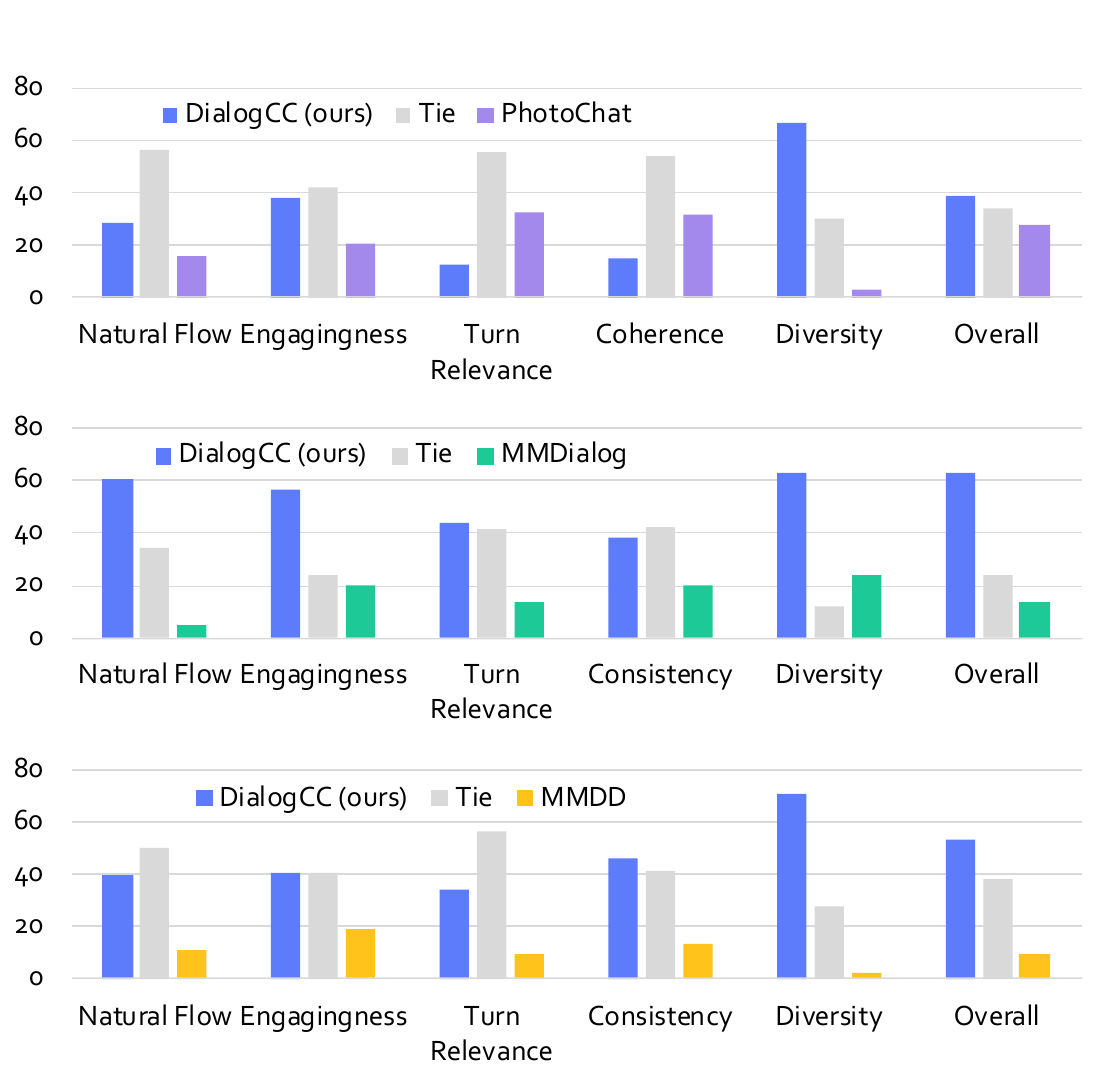}

\caption{Results of head-to-head comparison between DialogCC (ours) and three existing datasets: PhotoChat, MMDialog, MMDD.}

\label{main_fig:head_to_head}
\vspace{-1em}
\end{figure}

\subsection{Analysis of DialogCC} \label{sec:dialogCC}

\paragraph{High-Quality.} 
To assess the quality of DialogCC, we conduct the human evaluations based on five criteria: (1) image-sharing turn relevance, (2) image-sharing speaker adequacy, (3) image-sharing rationale relevance, (4) aligned image relevance, and (5) image consistency.
Each human rates 250 randomly chosen samples using a 4-point Likert scale for all criteria, except for (2) (\ie ``Yes'' or ``No''). Further details are in Appendix~\ref{supp_sec:human_rating}.
On average, we achieve higher scores across all evaluation criteria: 3.68 for (1), 95.1\% (``Yes'' ratio) for (2), 3.41 for (3), 3.30 for (4), and 3.57 for (5). 
In addition, we measure the inter-rater agreement using Krippendorff’s $\alpha$. On average, we get 0.39, which indicates fair agreement.
These results underscore the efficacy of our fully automatic pipeline, leveraging GPT-4 and CLIP. Breakdown analysis and details of human evaluation are shown in Appendix~\ref{supp_sec:detail_human_eval}.

To assess the quality gap between DialogCC and real-world scenarios, we conduct head-to-head human evaluations by comparing DialogCC with MMDD~\cite{lee2021constructing}, PhotoChat~\cite{zang2021photochat}, and MMDialog~\cite{feng2022mmdialog}.
We randomly sample 100 dialogues from each dataset and evaluate them based on six criteria: (1) natural flow, (2) engagingness, (3) turn relevance, (4) context consistency, (5) diversity, and (6) overall. Further details are in Appendix~\ref{supp_sec:head_to_head}.
As shown in Figure~\ref{main_fig:head_to_head}, DialogCC achieves a higher score in overall quality, particularly surpassing MMDialog by a large margin.
Furthermore, due to the nature of social media, MMDialog lacks natural conversational flow and engagingness compared to DialogCC by a large margin. 
This implies that while social media-sourced datasets may have significant advantages in terms of scale (in Table~\ref{main_tab:dataset_stat}), their quality is not guaranteed for the social dialogue domain.
Interestingly, compared to the PhotoChat, humans predominantly choose ``Tie''.
This indicates that although DialogCC is built fully automatically, its quality closely matches human-authored datasets.
Compared to the MMDD, DialogCC has more consistency between aligned images and dialogue context because we generate contextual image descriptions by prompting GPT-4.

{\renewcommand{\arraystretch}{1.35}
\begin{table}[t]
\centering
\begin{adjustbox}{width=0.9\columnwidth}

\begin{tabular}{@{}lccccc@{}}
\toprule
Dataset                                & \makecell{\# Unique \\ Dialog}      & \makecell{\# Unique \\Image}    & \makecell{Avg. \\U./D.}       & \makecell{Avg. \\I./D.} & \makecell{Avg. \\I./U.} \\ \midrule
PhotoChat     
                                  & 11,820          & 10,479           & 12.74         & 1.00          & 1.00          \\  
MMDD       
                                  & 17,679          & 13,288           & 11.56         & 1.76          & 1.29          \\  
MMDialog
                                  & 1,079,117       & 1,556,868        & 4.56          & 2.82          & 1.36          \\  
\textbf{DialogCC (ours)} 
                                 & \textbf{83,209} & \textbf{129,802} & \textbf{8.20} & \textbf{7.34} & \textbf{4.77} \\  \bottomrule                       
\end{tabular}

\end{adjustbox}
\caption{In total, DialogCC includes the
largest number of Avg. I./D. and I./U. than others. I./D.
and I./U. denote images by dialogue and images by an utterance, respectively. U./D. denotes utterances by a dialogue. More detailed statistics are in the Table~\ref{supp_tab:full_dataset_stat}.}
\label{main_tab:dataset_stat}

\end{table}}

\begin{figure}[t]
    \centering
    \includegraphics[width=\columnwidth]{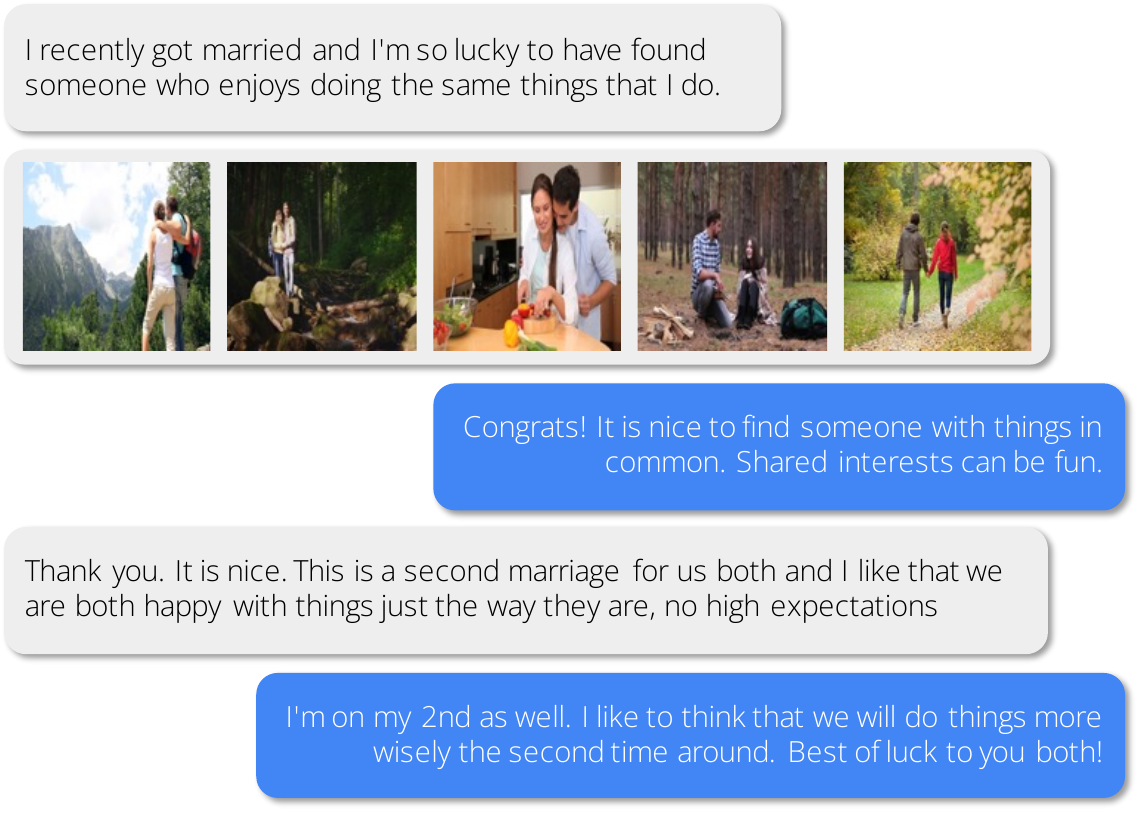}
    \caption{We present an example of DialogCC. More examples are in Appendix~\ref{supp_sec:more_examples}. Note that during actual model training, one of these images is randomly sampled to enhance the model's generalization capability.}
    \label{main_fig:dialogcc_example}
\end{figure}

\paragraph{Image Diversity.}
In real-life scenarios, people can share images with different styles, views, or objects for the same dialogue and context. 
However, as shown in Table~\ref{main_tab:dataset_stat}, the existing datasets include few images per dialogue and image-sharing turn. 
This does not reflect real-life scenarios and can cause an overfitting problem by forcing a model to memorize the pairs of images and dialogues. 
To handle this problem, our dataset has many and various images per dialogue and image-sharing turn, which is shown in Figure~\ref{main_fig:dialogcc_example}. 
In DialogCC, there are an average of 7.34 images per dialogue and 4.77 images per image-sharing turn, leading to enhanced generalization performance (in Section~\ref{sub_sec:main_results}).

\begin{comment}
    \paragraph{Rationale Distribution.}

To gain a better understanding of the generated rationales, we conduct an analysis of their verb-noun patterns. We parse the rationales using spaCy~\cite{honnibal2020spacy} and extract the root verb along with its first direct noun object.
Since we constrain a rationale to start with ``To'' in the prompt, we only consider rationales with a ``To verb noun'' structure during this analysis.
Out of a total of 106,063 generated rationales, 102,554 rationales follow this structure, whereas 3,509 rationales contain more complex clauses (e.g., \textit{To show how he spent his relaxing weekend.}).

In this analysis, we observe that the verb ``provide'' is used most frequently. This indicates that image sharing is intended to provide additional information related to the context of the dialogue. The tendency to provide additional information through image sharing is also evident in the verbs ``show'' and ``share''.

\input{tables/main_tables/rationale_distribution}
\end{comment}

\section{Experimentals} \label{sec:experiment_setting}

To explore how our dataset affects both text and image retrieval tasks, we implement two simple and standard baseline retrieval models for text-to-image and image-to-text settings. 

\subsection{Task Definition}

Follwing~\cite{lee2021constructing,zang2021photochat}, we explain the formulation of two main tasks - next response prediction and image retrieval. 
Let us assume that we have a multi-modal dialogue $\mathcal{D} = \{(u_j, i_j, c_j)\}_1^N$ where $N$ denotes the number of dialogue turns, and $j=t$ is the turn that an image sharing behavior occurs. Then, each task is formulated as follows.
\textbf{(1) Next response prediction} is to predict the next utterance at turn $t+1$ given the dialogue history ($\{u_j\}_1^t$) and image $i_t$.
\textbf{(2) Image retrieval} is to retrieve relevant image at turn $t$ given the dialogue history ($\{u_j\}_1^{t-1}$).
Following~\cite{shuster2018image,lee2021constructing}, we set the the number of retrieval candidates to 100 and use Recall@\{1,5,10\} and mean reciprocal rank (MRR) for the evaluation metrics.

\subsection{Datasets}

\textbf{(1) DialogCC (ours)} is a high-quality and diverse multi-modal dialogue dataset created by our proposed automatic pipeline powered by GPT-4 and CLIP models, which is described in Section~\ref{sec:method}. 
\textbf{(2) MMDD}~\cite{lee2021constructing} contains 45k multi-modal dialogues, where each utterance is replaced into a relevant image matched by their automatic pipeline.
\textbf{(3) PhotoChat}~\cite{zang2021photochat} contains 10k multi-modal dialogues, where the dialogue is constructed via a crowd-sourcing platform. 
\textbf{(4) MMDialog}~\cite{feng2022mmdialog} contains 1M multi-modal dialogues, where the dialogue is obtained from the social media platform.

\subsection{Baseline Models} \label{subsubsec:baseline}

The following are brief descriptions of two baseline retrieval models; more detailed information is provided in Appendix~\ref{supp_sec:baselines}. Two baseline models have a dual-encoder structure which consists of text encoder and image encoder. For the text encoder, we use the BERT-base~\cite{devlin2018bert} architecture (12 layers, 12 attention heads, 768 dimensions, uncased version). For the image encoder, we use the CLIP-B/32~\cite{radford2021learning} model. 

\subsection{Main Results} \label{sub_sec:main_results}

\begin{figure}[t]
    \centering
    \includegraphics[width=\columnwidth]{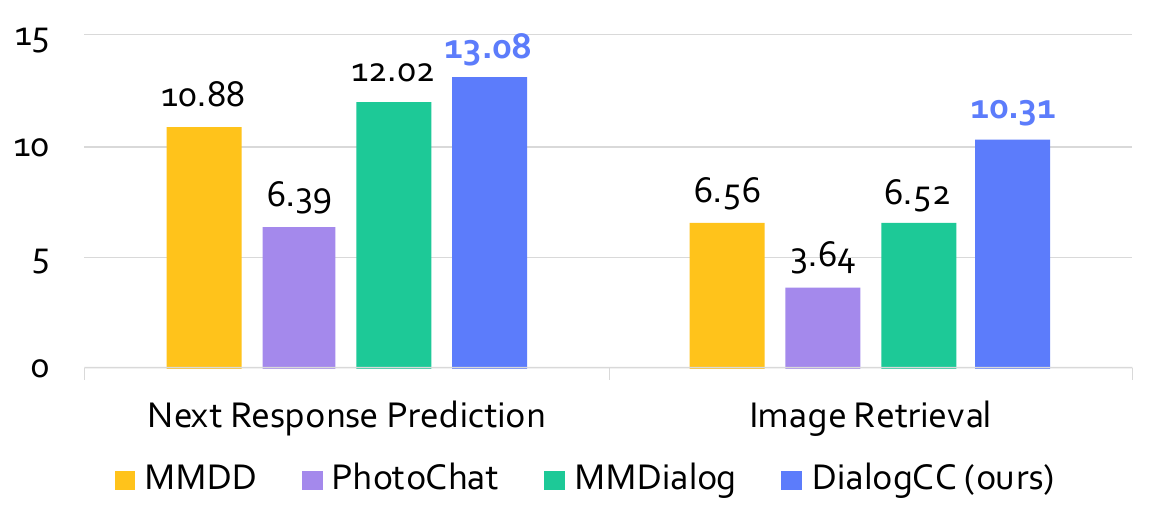}
    \caption{We report the average contributed performance on both tasks. Full results are in Appendix~\ref{supp_sec:full_rc_results}.}
    \label{main_fig:main_result}
    %\vspace{-1em}
\end{figure}

\paragraph{DialogCC contributes to the model's robustness.}
To understand the contributed impact of DialogCC on other dialogue datasets, we evaluate the baseline models trained on DialogCC on unseen dialogue datasets, MMDD, PhotoChat, and MMDialog. In other words, we differentiate training and evaluation datasets to observe how much each dataset can boost the model's generalization performance on unseen dialogue scenarios in the next response prediction task and the image retrieval task.
Figure~\ref{main_fig:main_result} summarizes the average contributed performance of each datasets. Although the scale of DialogCC is significantly smaller than MMDialog (83K vs. 1M), DialogCC contributes to the model's understanding of the unseen dialogue dataset on both tasks. This suggests that increasing the quality of the dataset is more important than the scale, which is in line with the direction of recent studies~\cite{zhou2023lima,xu2023demystifying} in data-centric AI. 

In addition, PhotoChat, which is manually constructed, underperforms compared to other datasets as indicated by its limited diversity (see Table~\ref{main_tab:dataset_stat}). This finding implies that, although our pipeline is automated compared to dialogue crowdsourcing, it not only ensures quality but is also more time and cost-efficient. This aligns with recent studies~\cite{lee2022personachatgen,kim2022soda} that generate dialogue datasets with the use of large language models, such as ChatGPT.

{\renewcommand{\arraystretch}{1.35}
\begin{table}[t]
\centering
\begin{adjustbox}{width=\columnwidth}
\begin{tabular}{@{}lcccccccc@{}}
\toprule
                         & \multicolumn{4}{c}{Image-Chat}                                     & \multicolumn{4}{c}{MPChat}                                        \\ \cmidrule(l){2-9} 
Train Dataset            & R@1            & R@5            & R@10           & MRR            & R@1            & R@5            & R@10           & MRR            \\ \midrule
MMDD                     & 12.66          & 31.35          & 43.02          & 22.86          & 13.41          & 36.05          & 53.33          & 25.72          \\
PhotoChat                & 6.89           & 21.41          & 32.54          & 15.76          & 7.52           & 29.23          & 43.57          & 19.15          \\
MMDialog                 & 14.37          & 32.45          & 43.39          & 24.29          & 21.94          & 51.01          & 66.90          & 36.08          \\
\textbf{DialogCC (ours)} & \textbf{20.22} & \textbf{42.60} & \textbf{54.86} & \textbf{31.65} & \textbf{29.77} & \textbf{57.91} & \textbf{70.70} & \textbf{42.84} \\ \bottomrule

\end{tabular}
\end{adjustbox}
\caption{We report the next response prediction performance on Image-Chat~\cite{shuster2018image} and MPChat~\cite{ahn2023mpchat} following the same evaluation setting.}
\label{main_tab:out_of_domain}
\vspace{-1em}

\end{table}
}

\paragraph{DialogCC improves the comprehension of the interaction between dialogue and images.} We evaluate the baseline models on two unseen multi-modal dialogue datasets, Image-Chat~\cite{shuster2018image} and MPChat~\cite{ahn2023mpchat}, which belong to the image-grounded dialogue dataset. Table~\ref{main_tab:out_of_domain} summarizes the zero-shot results of next response prediction task of models trained on four different datasets: MMDD, PhotoChat, MMDialog, and DialogCC. 
The model trained on DialogCC outperforms those trained on other datasets. This indicates that DialogCC significantly improves the model's comprehension of the interaction between dialogue and images, even when the image-grounded dialogue datasets encompass various patterns in multi-modal dialogue scenarios. This improvement is attributed to DialogCC's high-quality and diverse images, as shown in Figure~\ref{main_fig:head_to_head}, underscoring the reliability of our pipeline.

{\renewcommand{\arraystretch}{1.35}
\begin{table}[t]
\centering
\begin{adjustbox}{width=0.7\columnwidth}
\begin{tabular}{@{}lcccc@{}}
\toprule
Model Inputs     & R@1   & R@5   & R@10  & MRR   \\ \midrule
Image Only       & 8.22  & 22.60 & 33.52 & 16.94 \\
Dialogue Only    & 34.41 & 65.36 & 77.37 & 48.67 \\
Dialogue + Image & \textbf{40.64} & \textbf{71.46} & \textbf{81.99} & \textbf{54.61} \\ \bottomrule
\end{tabular}
\end{adjustbox}
\caption{We show the effectiveness of image modality in DialogCC on the next response prediction task.}
\label{main_tab:abl_modality}
\end{table}
}

\paragraph{Our pipeline effectively aligns dialogue with images.} 
Since we align two distinct modalities -- dialogue and images -- using GPT-4 and CLIP automatically, we evaluate the model by varying the input modalities to investigate the correlation between dialogue and images. As shown in Table~\ref{main_tab:abl_modality}, providing only images to the model results in significantly lower performance, indicating the importance of dialogue context in multi-modal dialogue tasks. The model, when considering only dialogue, shows comparable performance, possibly because our dataset is based on original text-only dialogues. Notably, considering both dialogue and image leads to better performance. These results suggest that the image modality enhances the understanding of the dialogue without disrupting its flow, benefiting from our robust and reliable alignment process.

\begin{figure}[t]
    \centering
    \includegraphics[width=\columnwidth]{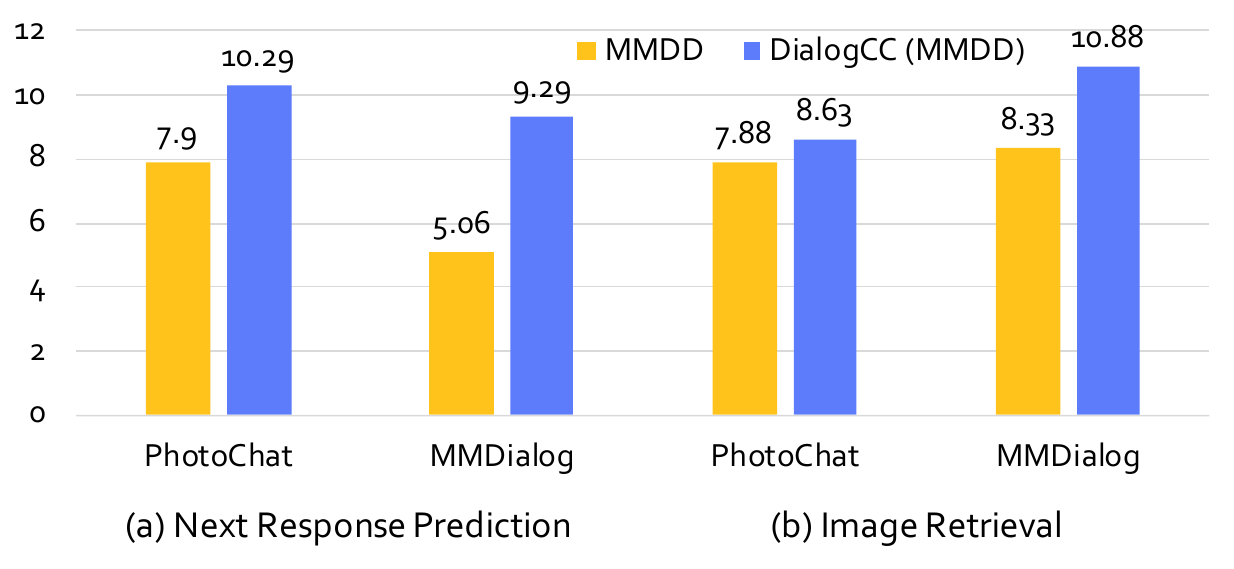}
    \caption{Results of the model trained on a sub-set of DialogCC using the same seed dialogue datasets as in MMDD.}
    \label{main_fig:mmdd_dialogcc}
    \vspace{-1em}
\end{figure}

\paragraph{Our pipeline is better than the semi-automatic method.}

To validate our automatic pipeline, we evaluate a model trained on a subset of DialogCC, using the same seed dialogue datasets as in MMDD (i.e., EmpatheticDialogues~\cite{rashkin2018towards}, Persona-Chat~\cite{zhang2018personalizing}, DailyDialog~\cite{li2017dailydialog}). Figure~\ref{main_fig:mmdd_dialogcc} demonstrates that, despite using identical dialogue datasets, our model significantly outperforms the one trained on MMDD. This result suggests that our pipeline more effectively discerns better image-sharing moments, taking advantage of GPT-4's capabilities. The performance gains are particularly notable in the MMDialog dataset, which includes a substantial number of images (see Table~\ref{main_tab:dataset_stat}). This enhancement can be attributed to the use of a diverse image dataset (\ie CC3M) as the seed dataset. Furthermore, our pipeline, requiring minimal human intervention, not only enhances the generalization performance of the trained model but also proves to be cost-effective, thereby ensuring both quality and performance.

\begin{figure}
    \centering
    \includegraphics[width=\columnwidth]{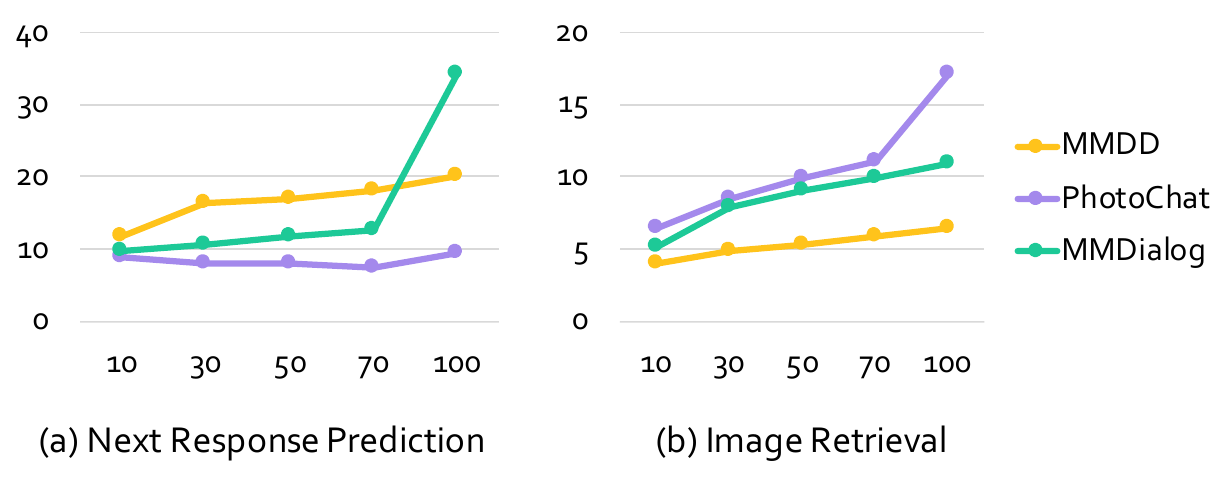}
    \caption{Scaling results of the model trained on DialogCC on both tasks.}
    \label{main_fig:scaling}
    \vspace{-1em}
\end{figure}

\paragraph{Our pipeline benefits from scaling up the dataset size.}
To investigate whether our pipeline benefits from dataset scaling, we evaluate the model trained on DialogCC with varying dataset sizes. As shown in Figure~\ref{main_fig:scaling}, increasing the dataset size significantly enhances performance on three previously unseen dialogue datasets. These results indicate that our pipeline indeed benefits from scaling up the dataset size, thereby ensuring the creation of reliable and high-quality datasets. In future work, we plan to construct a million-scale, multi-modal dialogue dataset using the SODA~\cite{kim2022soda} dataset in conjunction with our pipeline.

\subsection{Case Study} \label{sec:case_study}

\begin{figure*}[t]
    \centering
    \includegraphics[width=\textwidth]{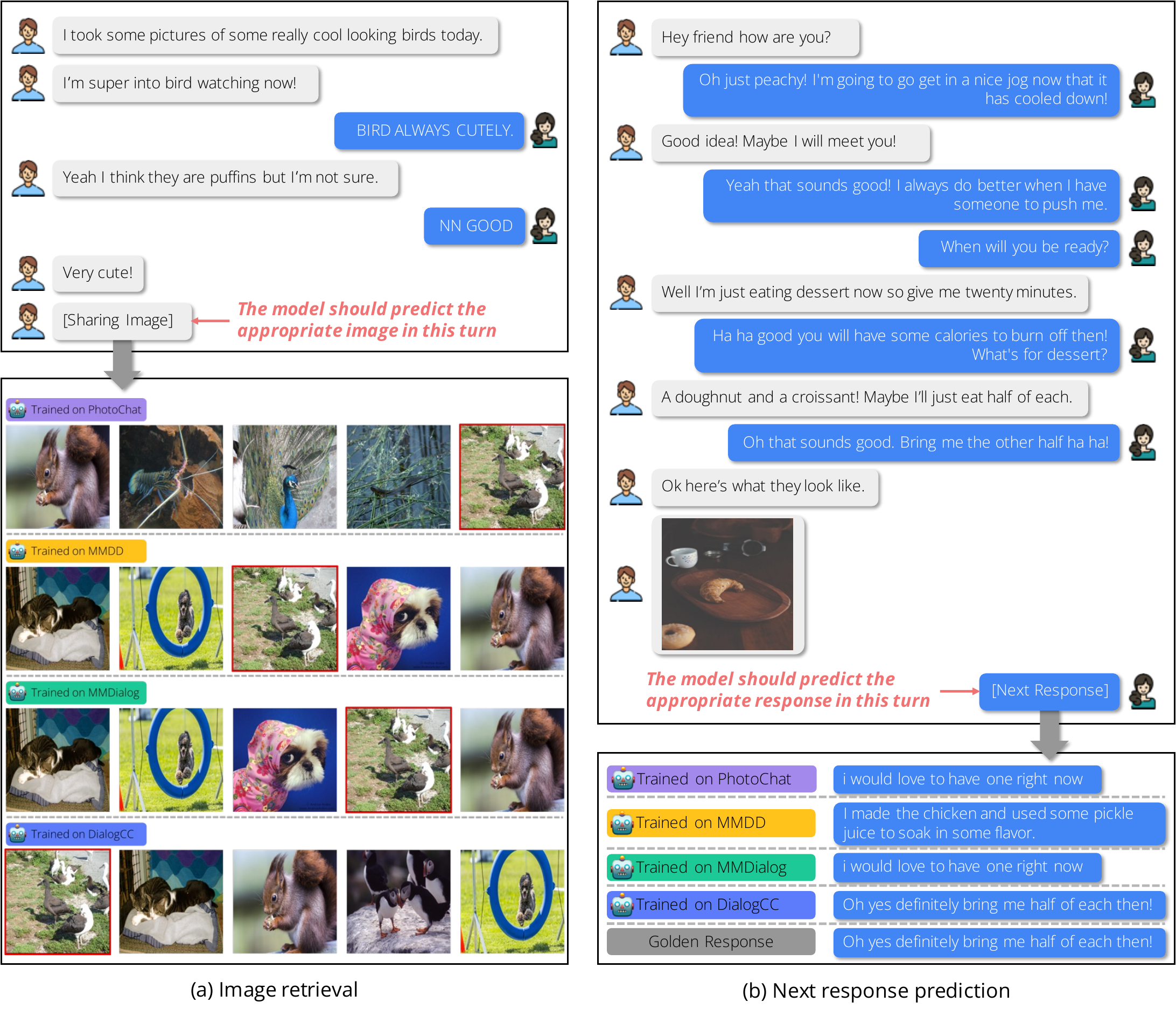}
    \caption{Two examples of retrieved results (\ie (a) image retrieval and (b) next response prediction) from models trained on four different datasets. Each provided dialogue is from the PhotoChat dataset. In (a), we display the top-5 ranked images from left to right, with the ground-truth image marked in red. In (b), only the top-1 ranked next response is shown. Note that neither the \texttt{[Sharing Image]} turn nor the \texttt{[Next Response]} turn is provided to the model's input during the inference stage. More examples are presented in Figure~\ref{supp_fig:case_study_dialogcc}.}
    \label{main_fig:case_study}
    \vspace{-1em}
\end{figure*}

As shown in Figure~\ref{main_fig:case_study}, we present two examples of results retrieved from models trained on four different datasets. In Figure~\ref{main_fig:case_study}-(a), three models trained on previous datasets (\ie PhotoChat, MMDD, MMDialog) retrieve inappropriate images by focusing on the word ``cute'' in the last utterance. Conversely, the model trained with DialogCC accurately retrieves a suitable image of the cute animal ``puffins.'' This indicates that the model trained on DialogCC not only recognizes what a ``puffin'' looks like but also understands the contextual relevance of the word ``cute'' within the entire dialogue. This capability is attributed to the high-quality and diverse imagery of our dataset. In Figure~\ref{main_fig:case_study}-(b), DialogCC significantly enhances the model's ability to understand multi-modal dialogues, resulting in the accurate retrieval of the correct subsequent response. These results highlight the importance of both high-quality and image diversity in developing a more generalized and robust model.

\section{Conclusion} \label{sec:conclusion}

In this paper, we propose the automatic pipeline for creating a multi-modal social dialogue dataset that involves aligning and filtering with GPT-4 and CLIP, respectively. We also propose a large-scale and high-quality multi-modal dialogue dataset, DialogCC, which is constructed by leveraging the automatic pipeline with five text-only dialogue datasets and an image-text pair CC3M dataset. 
In a comprehensive analysis, compared to existing datasets MMDD, PhotoChat, MMDialog, using DialogCC helps achieve better quality in terms of various metrics. 
Moreover, our dataset consists of many and various images per dialogue that can be beneficial in model generalization performance. Extensive experiments demonstrate that a model trained with DialogCC increase model's robustness.

\section*{Limitations}

\paragraph{Societal Impact.} 
As reported in~\cite{wang2021gender}, even if we give the gender-neutral query to CLIP~\cite{radford2021learning} model, the CLIP model sometimes retrieves images causing gender-bias issues. We are concerned that this problematic issue may exist in our dataset because we use the CLIP model to match relevant images to the generated image description by GPT-4. A notable example of this bias is the association of women's images with the profession of ``hair designer.'' Such biases are concerning as they could propagate stereotypes. Therefore, the image retrieval model trained on our dataset may sometimes retrieve biased images. We should consider this problem important when building a multi-modal search model. In the future work, we will mitigate this issue to be fairer and more generalized model.

\paragraph{Addressing Cross-Turn Image Inconsistency.}
In our effort to construct a natural and coherent multi-modal dialogue dataset, we utilize GPT-4 to identify appropriate moments for image sharing within text-only dialogues, ensuring conversational flow. We then generate image descriptions for these moments and align the corresponding images using CLIP. To maintain single-turn image consistency, we introduce a straightforward algorithm based on pairwise similarity comparisons through CLIP. Nevertheless, our approach currently overlooks cross-turn image inconsistencies within the same dialogue, and addressing this challenge is part of our future objectives.

\paragraph{Considering Personalization.}
Our dataset aims to enhance generalization performance by mapping multiple images to a single utterance. This dataset, while beneficial for model generalization, may occasionally result in the sharing of images unrelated to the speaker's specific subject, diminishing user interactability. For example, if a speaker refers to their Chihuahua, the model might incorrectly present an image of a Golden Retriever due to the broad mapping in our dataset. Recognizing these limitations, we emphasize the importance of not only improving generalization but also incorporating user preferences to bolster engagement. Our future work is thus dedicated to developing a personalized multi-modal dialogue dataset and system.

\paragraph{Improving Factuality in Alignment.}
Despite our meticulous efforts in developing DialogCC with carefully designed pipelines, the dataset may still include samples that are factually inaccurate. For example, an image meant to illustrate an utterance related to ``race walking'' might instead show a ``marathon scene,'' or an utterance describing a ``three-story hotel building'' could be incorrectly matched with a photo of a ``four-story hotel.'' In the future, we will consider real scene understanding~\cite{lee2024collavo, lee2024moai} to enhance the factual accuracy of the alignment process.

\section*{Acknowledgement}

This work was supported by Institute of Information \& communications Technology Planning \& Evaluation (IITP) grant funded by the Korea government(MSIT) [No.2022-0-00641, XVoice: Multi-Modal Voice Meta Learning]

\bibliography{anthology,custom}

\clearpage

\appendix

\section{Details of Source Datasets} \label{supp_sec:src_dataset}

\subsection{Source Dialogue Datasets} \label{supp_sec:src_dialogue}

We collect the give text-only social dialogue datasets (i.e., Wizard-of-Wikipedia~\cite{dinan2018wizard}, Persona-Chat~\cite{zhang2018personalizing}, EmpatheticDialogues~\cite{rashkin2018towards}, DailyDialog~\cite{li2017dailydialog}, and BlendedSkillTalk~\cite{smith2020can}) through the ParlAI~\cite{miller2017parlai} framework, which provides many dialogue datasets online. The statistics of source dialogue datasets are shown in Table~\ref{supp_tab:source_dataset_stat}. The details of each dataset are described as follows:

\paragraph{Wizard-of-Wikipedia.} 
This dataset aims to enable the dialogue agent to generate knowledgeable responses grounded in information retrieved from Wikipedia to enhance the engagement of the conversation. The dataset was constructed via a crowdsourcing platform, where two participants converse with each other on one of a total of 1,365 topics. One participant selects a conversational topic and assumes the role of a knowledgeable expert (referred to as the \textit{wizard}), while the other acts as a curious learner (the \textit{apprentice}). The dataset can be downloaded from the ParlAI framework by setting the task name as \texttt{\small wizard\_of\_wikipedia:basic\_apprentice\_dialog}.

\paragraph{Persona-Chat.} 
This dataset is designed to enable the dialogue agent to generate responses based on personal information, whether their own or others. It was constructed using a crowdsourcing platform, where two participants engage in a conversation based on provided persona information. The persona is represented by a set of sentences that depict demographic and psychographic characteristics~\cite{lee2022personachatgen}. Examples of such sentences include ``I am getting old.'' and ``I love the color blue.'' Given that the original persona sentences exhibit simple linguistic structures, a revised version of these sentences is also provided to make the model training more challenging and thereby enhance performance. To download this dataset from ParlAI, set the task name to \texttt{personachat:both\_original}.

{\renewcommand{\arraystretch}{1.35}
\begin{table}[t]
\centering
\begin{adjustbox}{width=\columnwidth}

\begin{tabular}{lccccc} 
\toprule
Dataset                                      & Type  & \# Dialog & \# Utter & \makecell{Avg. \\ Utter. Len} & \makecell{Avg. \\ Utter/Dialog} \\ \midrule
\multirow{4}{*}{Blended Skill Talk}        & train & 4,819       & 54,036       & 13.09           & 11.21             \\ 
                                             & valid & 1,009       & 11,302       & 13.17           & 11.20             \\
                                             & test  & 980         & 10,964       & 13.60           & 11.19             \\ \cmidrule{2-6}
                                             & total & 6,808       & 76,302       & 13.29           & 11.20             \\ \midrule
\multirow{4}{*}{DailyDialog}       & train & 21,753      & 152,104      & 11.44           & 6.99              \\
                                             & valid & 1,960       & 14,138       & 11.36           & 7.21              \\
                                             & test  & 1,958       & 13,480       & 11.56           & 6.88              \\
                                             \cmidrule{2-6}
                                             & total & 25,671      & 179,722      & 11.45           & 7.03              \\
                                             \midrule
\multirow{4}{*}{EmpatheticDialogues}       & train & 19,531      & 80,508       & 13.45           & 4.12              \\
                                             & valid & 2,769       & 11,476       & 14.50           & 4.14              \\
                                             & test  & 2,547       & 10,518       & 15.33           & 4.13              \\
                                             \cmidrule{2-6}
                                             & total & 24,847      & 102,502      & 14.43           & 4.13              \\
                                             \midrule
\multirow{4}{*}{Persona-Chat}  & train & 8,939       & 131,438      & 10.09           & 14.70             \\
                                             & valid & 1,000       & 15,602       & 10.30           & 15.60             \\
                                             & test  & 968         & 15,024       & 10.19           & 15.52             \\
                                             \cmidrule{2-6}
                                             & total & 10,907      & 162,064      & 10.19           & 15.28             \\
                                             \midrule
\multirow{4}{*}{Wizard of Wikipedia} & train & 18,430      & 166,787      & 16.37           & 9.05              \\
                                             & valid & 1,948       & 17,715       & 16.40           & 9.09              \\
                                             & test  & 1,933       & 17,497       & 16.26           & 9.05              \\
                                             \cmidrule{2-6}
                                             & total & 22,311      & 201,999      & 16.34           & 9.07     \\  \bottomrule                       
\end{tabular}

\end{adjustbox}
\caption{We show the statistics of source dialogue datasets.}
\label{supp_tab:source_dataset_stat}

\end{table}}

\setlength{\columnsep}{0.2cm}
\begin{figure*}[!t]
%\setlength{\columnsep}{0.2cm}

%\vspace{3mm}
%\begin{multicols}{2}
\begin{tcolorbox}[
    colback=qualcolor!5!white,
    colframe=qualcolor!75!black,]
\begin{small}
\textbf{Prompt Template for Inferring Image-Sharing Moments:}\\
The following is a dialogue between \texttt{[speaker1]} and \texttt{[speaker2]}. The dialogue is provided line-by-line. In the given dialogue, select all utterances that are appropriate for sharing the image in the next turn, and write the speaker who will share the image after the selected utterance. You should also provide a rationale for your decision and describe the relevant image concisely.\\
\\
Dialogue:\\
\texttt{[dialogue]} \\
\\
Restrictions: \\
(1) your answer should be in the format of ``<UTTERANCE> | <SPEAKER> | <RATIONALE> | <IMAGE DESCRIPTION>''. \\
(2) you MUST select the utterance in the given dialogue, NOT generate a new utterance. \\
(3) the rationale should be written starting with "To". \\
\\
Answer:\\
1.\\
\end{small}
%\end{tcolorbox}
\Sepline

%\begin{tcolorbox}[colback=qualcolor!5!white,colframe=qualcolor!75!black]
\begin{small}
\textbf{Prompt Template for Identifying Speaker Names in \texttt{[dialogue]}:}\\
\texttt{[dialogue]} \\
\\
Q: What are the names of Speaker A and Speaker B in the given dialogue? Your answer should be in the format of "<Speaker A> | <Speaker B>".\\
A:\\
\end{small}

\end{tcolorbox}
%\end{multicols}

\caption{A prompt template for inferring image-sharing moments (\textbf{top}). A prompt template for identifying speaker names in \texttt{[dialogue]} (\textbf{bottom}).}
\label{supp_fig:prompt_template}
\end{figure*}

\paragraph{EmpatheticDialogues.} 
This dataset is designed to enable dialogue agents to generate empathetic responses by understanding and interpreting the interlocutor's emotional situation. It was constructed using a crowdsourcing platform where two turkers are assigned specific roles: \textit{speaker} and \textit{listener}. The speaker is provided with an emotional situation and one emotion label from a set of 32 labels, while the listener responds with empathy to the speaker's situation. The dataset can be downloaded from the ParlAI framework using the task name \texttt{empathetic\_dialogues}.

\paragraph{DailyDialog.} 
This dataset was constructed by crawling daily-life conversations from various websites. It includes additional information crucial for understanding and proceeding with daily-life conversations between partners, such as emotion, topic, and dialog act. Specifically, there are seven emotion categories: anger, disgust, fear, happiness, sadness, surprise, and others. The dataset contains 10 daily topics: ordinary life, school life, culture \& education, attitude \& emotion, relationship, tourism, health, work, politics, and finance. Additionally, there are four dialog acts: inform, question, directive, and commission. The dataset can be downloaded from the ParlAI framework using the task name \texttt{dailydialog:no\_start}.

\paragraph{Blended Skill Talk.} 
This dataset is designed to help dialogue agents learn how to use multiple conversational skills interactively and naturally rather than relying on a single isolated skill. The dataset was constructed by integrating several skills (\ie empathetic, knowledgeable, and personalizing) into a single conversation via a crowdsourcing platform. Within this dataset, there are four skill annotations: (1) Knowledge, (2) Empathy, (3) Personal situations, and (4) Personal background. Each utterance in a conversation is annotated with a corresponding skill. The dataset can be downloaded from the ParlAI framework using the task name \texttt{blended\_skill\_talk}.

\subsection{Source Image-Caption Pair Dataset}

We download the Conceptual Captions 3M~\cite{sharma2018conceptual} (CC3M) dataset in here~\footnote{\url{https://ai.google.com/research/ConceptualCaptions/download}}. Since the CC3M dataset provides image URLs, we download images using img2dataset~\footnote{\url{https://github.com/rom1504/img2dataset}} library, which is a helpful library for quick downloading large-scale images based on URLs. 
We downloaded images in March 2023 and we store downloaded images as a \texttt{jpg} format. We obtain 2,783,547 images from the train set and 12,911 from the valid set. Note that because each image URL has the copyright, we only use opened URLs as source image-caption data when we create DialogCC.

\subsection{Licenses}

We list the licenses of each source dataset that we utilized in the creation of DialogCC.

\begin{itemize}
    \item Wizard-of-Wikipedia: CC-BY-4.0
    \item Persona-Chat: CC-BY-4.0
    \item EmpatheticDialogues: CC-BY-4.0
    \item DailyDialog: CC BY-NC-SA 4.0
    \item Blended Skill Talk: CC-BY-4.0
    \item Conceptual Caption 3M: Open License by Google
\end{itemize}

CC3M is under the Google open license, which allows for the free use of the dataset for any purpose. Since all the datasets except DailyDialog are permissible for commercial use, we will release our dataset DialogCC by following the ``C BY-NC-SA 4.0'' license. This means the dataset can only be used for academic or research purposes and is not permitted for commercial use.

\section{Details of Automated Pipeline} \label{supp_sec:pipeline}

\subsection{Prompt Templates} \label{supp_sec:prompt_template}

In order to infer image-sharing moments using GPT-4, we thoughtfully create the prompt template, as depicted in Figure~\ref{supp_fig:prompt_template}.
We provide GPT-4 with specific guidelines (\ie \textit{restrictions}) derived from insights gained in a preliminary study to ensure the generation of higher-quality results.
Specifically, the model produces potential image-sharing utterances with speaker (\textit{who}), rationale (\textit{why}), and image description (\textit{what}). 
Moreover, regarding the second sentence in the restrictions, if we omit this from the prompt, the model occasionally fails to infer the image-sharing utterance within the given dialogue. 
Instead, it creates a new utterance suggesting an event that might occur following the current dialogue context.
For the \texttt{[dialogue]}, we provide the entire dialogue history into the model. The motivation behind this design decision is explained in Section~\ref{supp_sec:full_history}.
Furthermore, as we mentioned in Section 3.2, to make the \texttt{[dialogue]} natural, we identify the actual speaker names within the given \texttt{[dialogue]} based on the designed prompt template as shown in Figure~\ref{supp_fig:prompt_template}.
To parse the \textit{utterance}, \textit{speaker}, \textit{rationale}, and \textit{image description} from the GPT-4 generation results, we implemented a simple parser using regex patterns, as depicted in Figure~\ref{supp_fig:parser}.
%\texttt{$r'^(?:\d+\.\s+)?\"?(?P<utterance>.*?)\"?\s+\|\s+(?P<speaker2>.*?)(?:\s+\|\s+(?P<rationale>.*?))?(?:\s+\|\s+(?P<description>.*?))?$'$}

\begin{figure*}[t]
\begin{lstlisting}[language=Python, basicstyle=\ttfamily\small,keywordstyle = {\bfseries \color[cmyk]{0,1,0,0}},stringstyle = {\ttfamily \color[rgb]{0,0,1}},
breaklines = true,
breakindent = 10pt,
commentstyle = {\itshape \color[cmyk]{1,0.4,1,0}},]
import re
from typing import Dict

class Parser:

    PATTERN = r'^(?:\d+\.\s+)?\"?(?P<utterance>.*?)\"?\s+\|\s+(?P<speaker>.*?)(?:\s+\|\s+(?P<rationale>.*?))?(?:\s+\|\s+(?P<description>.*?))?$'
    
    def parse(pred: str) -> Dict:
        pred = pred.strip()
        
        matches = re.finditer(Parser.PATTERN, pred, re.MULTILINE)
        results = []
        for match in matches:
            utter = match.group('utterance')
            speaker = match.group('speaker') 
            rationale = match.group('rationale')
            description = match.group('description')
            
            results.append({
                'utterance': utter,
                'speaker': speaker,
                'rationale': rationale,
                'description': description
            })

        return results
\end{lstlisting}
\caption{A Python code for parsing generated responses from GPT-4.}
\label{supp_fig:parser}
\end{figure*}

{\renewcommand{\arraystretch}{1.35}
\begin{table}[t]
\centering
\begin{adjustbox}{width=\columnwidth}

\begin{tabular}{cccccc} \toprule
           & Similarity & Q1     & Q2     & Q3     & Q4     \\ \midrule
Similarity &     \graycell       & 0.3066 & 0.3153 & 0.3573 & 0.2478 \\
Q1         &    \graycell        &    \graycell    & 0.5566 & 0.4496 & 0.3313 \\
Q2         &      \graycell      &    \graycell    &    \graycell    & 0.7999 & 0.4461 \\
Q3         &     \graycell       &     \graycell   &    \graycell    &     \graycell   & 0.5826 \\
Q4         &       \graycell     &    \graycell    &    \graycell    &  \graycell      &       \graycell                               
 \\  \bottomrule                       
\end{tabular}

\end{adjustbox}
\caption{We show Spearman's correlation between four human evaluation items and utterance-image cosine similarity using CLIP ViT-L/14 model. Q1, Q2, Q3, and Q4 denote the turn relevance, rationale relevance, aligned image relevance, and image consistency, respectively.}
\label{supp_tab:correlation}

\end{table}}

\subsection{Motivation behind Providing Full Dialogue} \label{supp_sec:full_history}

The objective of this paper is to create a high-quality multi-modal dialogue dataset, building upon an existing text-only social dialogue dataset, as described in Section~\ref{supp_sec:src_dialogue}.
This implies that the source dialogue datasets already possess an inherent dialogue context, such as conversational flow, holistic meaning, and topic.
Therefore, it's imperative to identify potential image-sharing moments without disturbing the established conversational flow, even after integrating relevant images into the inferred image-sharing utterances. As a result, we feed the complete dialogue history to the model.

\subsection{Motivation behind Using GPT-4} \label{supp_sec:gpt-4}

The motivation behind using GPT-4 is to generate contextualized image descriptions rather than relying on the calculation of cosine similarity between a single utterance and an image, as done by image-text matching models (\eg VSRN) in MMDD.
Given that the image-text matching model is trained on image-caption pair datasets, it struggles to capture the holistic meaning from the dialogue context. 
For instance, it becomes challenging to identify relevant images for the sentence ``I ate it yesterday. See this photo!'' without access to the preceding dialogue context.
Furthermore, as depicted in Table~\ref{supp_tab:correlation}, the correlation between the CLIP similarity and the relevance of turns as rated by humans is low. 
This finding suggests that determining relevant images using only utterances is not effective.
Therefore, inspired by the recent advancements of large language models in the social dialogue domain~\cite{lee2022does,lee2022personachatgen,kim2022soda}, we choose to employ GPT-4 to generate contextualized image descriptions.

{\renewcommand{\arraystretch}{1.35}
\begin{table}[t]
\centering
\begin{adjustbox}{width=0.9\columnwidth}

\begin{tabular}{@{}lc@{}}
\toprule
Model            & Recall ($\uparrow$) \\ \midrule
Tulu-13B~\cite{wang2023far}         & 2.27   \\
WizardLM-13B~\cite{xu2023wizardlm}     & 18.80  \\
Vicuna-13B~\cite{chiang2023vicuna}       & 29.96  \\
LLaMA-2-Chat-13B~\cite{touvron2023llama} & 29.44  \\
\textbf{GPT-4}~\cite{openai2023gpt}            & \textbf{35.23}  \\ \bottomrule                    
\end{tabular}

\end{adjustbox}
\caption{We present a comparison of open-source LLMs, including GPT-4, based on the recall metric using the PhotoChat dataset.}
\label{supp_tab:open-source-llm}

\end{table}}

\subsection{GPT-4 versus Open-Sourced LLM} \label{supp_sec:open-source-llm}

GPT-4 can be replaced by open-sourced LLMs in our pipeline for cost reduction, leading to enhanced scalability of the dataset. To see their feasibility, we evaluate the LLM’s ability to infer image-sharing moments in PhotoChat, using recall as the metric, measuring whether one of the generated imagesharing
turns matches the ground-truth turn in PhotoChat. Table~\ref{supp_tab:open-source-llm} shows that GPT-4 outperforms recent open-sourced LLMs. Thus, we used GPT-4 since our work focuses on the quality and diversity of the multi-modal dialogue dataset. However, it is possible to create a large-scale dataset using our automatic pipeline with GPT-4. With adequate budgets, we can increase the dataset size considerably, ensuring quality and diversity.

\subsection{Details of Filtering Step}

We determine the threshold scores used in the \textit{filtering} step by manually evaluating randomly chosen 10,000 samples.

\section{Further Analyses on DialogCC}

{\renewcommand{\arraystretch}{1.35}
\begin{table}[t]
\centering
\begin{adjustbox}{width=\columnwidth}

\begin{tabular}{lcccc}
\toprule
Dataset         & \makecell{Source \\ Dialog}           & \makecell{Source \\ Image}         & \makecell{Interaction \\ Type} & \makecell{Aligning \\ Two Modalities} \\ \midrule
VisualDialog    & CS                      & COCO                  & grounding        & Human                                       \\ 
IGC             & CS                      & VQG                   & grounding        & Human                                       \\ 
ImageChat       & CS                      & YFCC100M              & grounding        & Human                                       \\ 
OpenViDial      & Movie \& TV & Movie \& TV          & grounding        & Human                                       \\ 
MMChat          & Social Media & Social Media       & grounding        & Human                                       \\ 
MPChat          & Reddit & Reddit           & grounding        & Human                                       \\ 
PhotoChat       & CS                      & \makecell{Open Image \\ Dataset V4} & sharing          & Human                                       \\  
MMDD            & ED, PC, Daily           & \makecell{MS-COCO, \\ Flicker 30k}  & sharing          & VSRN + Human                                \\  
MMDialog        & \multicolumn{2}{c}{Social Media}                & sharing          & Human                                       \\ \midrule
\bluecell DialogCC (ours) & \bluecell \makecell{ED, PC, \\ Daily, BST, WoW} & \bluecell CC3M                  & \bluecell sharing          & \bluecell GPT-4, CLIP                  \\  \bottomrule                       
\end{tabular}

\end{adjustbox}
\caption{Comparison of DialogCC with other multi-modal dialogue datasets: VisualDialog~\cite{das2017visual}, IGC~\cite{mostafazadeh2017image}, ImageChat~\cite{shuster2020multi}, OpenViDial~\cite{meng2020openvidial}, MMChat~\cite{zheng2021mmchat}, MPChat~\cite{ahn2023mpchat}, PhotoChat~\cite{zang2021photochat}, MMDD~\cite{lee2021constructing}, and MMDialog~\cite{feng2022mmdialog}. CS denotes crowdsourcing. ED, PC, Daily, BST, WoW denote EmpatheticDialogues, Persona-Chat, DailyDialog, BlendedSkillTalk, Wizard-of-Wikipedia. VSRN denotes the Visual Semantic Reasoning Network~\cite{li2019visual}.}
\label{supp_tab:dataset_comparison}

\end{table}}

\subsection{Comparing to Existing Multi-Modal Dialogue Datasets} \label{supp_sec:comparison}

Table~\ref{supp_tab:dataset_comparison} compares DialogCC with other multi-modal dialogue datasets. Unlike other image-grounded datasets, DialogCC falls under the category of image-sharing datasets in terms of multi-modal interaction type.
Specifically, image-grounded datasets always begin with a given image. Both conversational partners perceive this image and discuss it, such as questioning.
In other words, image-grounded datasets always start from the given image, then two conversational partners perceive the given image and then talk about the image. However, with image-sharing datasets, the two participants converse with each other before sharing an image. At some point, one of them shares a relevant image based on the preceding dialogue context. After this, the conversation continues, with both partners discussing the shared image. Thus, the image-sharing dialogue dataset is more challenging than image-grounded datasets, since the former encompasses the scope of the latter as well.

Among the existing image-sharing datasets, the alignment of two different modalities (\ie image and dialogue) is typically performed by humans. However, we leverage the GPT-4 and CLIP models to align these modalities without human intervention. Although DialogCC is fully constructed by automatic pipeline, it achieves high-quality and diverse alignments compared to other image-sharing datasets, as depicted in Figure 5.

{\renewcommand{\arraystretch}{1.35}
\begin{table*}[t]
\centering
\begin{adjustbox}{width=0.9\textwidth}

\begin{tabular}{lcccccccccc}
\toprule
Dataset                 & Type & \# Unique Dialog & \# Image          & \# Unique Image   & \# Utter          & Avg. U/D      & Avg. I/D      & \# Sharing Utter  & Avg. S/D      & Avg. I/S      \\ \midrule
\multirow{4}{*}{PhotoChat}       & train                    & 9,890           & 9,890            & 8,549            & 125,512          & 12.69         & 1.00          & 9,890            & 1.00          & 1.00          \\
                                 & valid                    & 962             & 962              & 962              & 12,205           & 12.69         & 1.00          & 962              & 1.00          & 1.00          \\
                                 & test                     & 968             & 968              & 968              & 12,421           & 12.83         & 1.00          & 968              & 1.00          & 1.00          \\ \cmidrule{2-11}
                                 & total                    & 11,820          & 11,820           & 10,479           & 150,138          & 12.74         & 1.00          & 11,820           & 1.00          & 1.00          \\ \midrule
\multirow{4}{*}{MMDD}            & train                    & 13,141          & 39,956           & 12,272           & 131,392          & 10.00         & 3.04          & 21,525           & 1.64          & 1.86          \\
                                 & valid                    & 2,148           & 2,401            & 334              & 26,576           & 12.37         & 1.12          & 2,401            & 1.12          & 1.00          \\
                                 & test                     & 2,390           & 2,673            & 682              & 29,453           & 12.32         & 1.12          & 2,673            & 1.12          & 1.00          \\ \cmidrule{2-11}
                                 & total                    & 17,679          & 45,030           & 13,288           & 187,421          & 11.56         & 1.76          & 26,599           & 1.29          & 1.29          \\ \midrule
\multirow{4}{*}{MMDialog}        & train                    & 1,059,117       & 2,981,568        & 1,509,284        & 4,825,053        & 4.56          & 2.82          & 2,193,816        & 2.07          & 1.36          \\
                                 & valid                    & 10,000          & 27,944           & 23,812           & 45,382           & 4.54          & 2.79          & 20,546           & 2.05          & 1.36          \\
                                 & test                     & 10,000          & 28,419           & 23,772           & 45,801           & 4.58          & 2.84          & 20,871           & 2.09          & 1.36          \\ \cmidrule{2-11}
                                 & total                    & 1,079,117       & 3,037,931        & 1,556,868        & 4,916,236        & 4.56          & 2.82          & 2,235,233        & 2.07          & 1.36          \\ \midrule
\multirow{4}{*}{DialogCC (ours)} & train                    & 68,269          & 699,505          & 101,877          & 552,991          & 8.10          & 10.25         & 106,063          & 1.55          & 6.60          \\
                                 & valid                    & 7,635           & 44,093           & 13,842           & 63,074           & 8.26          & 5.78          & 11,662           & 1.53          & 3.78          \\
                                 & test                     & 7,305           & 43,872           & 14,083           & 60,116           & 8.23          & 6.01          & 11,139           & 1.52          & 3.94          \\ \cmidrule{2-11}
                                 & \textbf{total}           & \textbf{83,209} & \textbf{787,470} & \textbf{129,802} & \textbf{676,181} & \textbf{8.20} & \textbf{7.34} & \textbf{128,864} & \textbf{1.54} & \textbf{4.77} \\  \bottomrule                       
\end{tabular}

\end{adjustbox}
\caption{In total, DialogCC includes the
largest number of Avg. I./D. and I./S. than others. I./D.
and I./S. denote images by dialogue and images by an image-sharing utterance, respectively. U./D. denotes utterances by a dialogue.}
\label{supp_tab:full_dataset_stat}

\end{table*}}

{\renewcommand{\arraystretch}{1.35}
\begin{table*}[t]
\centering
\begin{adjustbox}{width=0.9\textwidth}

\begin{tabular}{lcccccccccc}
\toprule
Dataset                 & Type & \# Unique Dialog & \# Image          & \# Unique Image   & \# Utter          & Avg. U/D      & Avg. I/D      & \# Sharing Utter  & Avg. S/D      & Avg. I/S      \\ \midrule
\multirow{4}{*}{BlendedCC}       & train                    & 4,595           & 52,890  & 25,916         & 51,650  & 11.24    & 11.51    & 7,671           & 1.67     & 6.89     \\
                             & valid                    & 927             & 6,185   & 4,047          & 10,376  & 11.19    & 6.67     & 1,458           & 1.57     & 4.24     \\
                             & test                     & 872             & 5,962   & 3,856          & 9,790   & 11.23    & 6.84     & 1,394           & 1.60     & 4.28     \\ \cmidrule{2-11}
                             & total                    & 6,394           & 65,037  & 33,819         & 71,816  & 11.22    & 8.34     & 10,523          & 1.61     & 5.14     \\ \midrule
\multirow{4}{*}{DailyCC}     & train                    & 19,459          & 162,260 & 42,088         & 139,416 & 7.16     & 8.34     & 26,495          & 1.36     & 6.12     \\
                             & valid                    & 1,665           & 7,322   & 3,644          & 12,228  & 7.34     & 4.40     & 2,248           & 1.35     & 3.26     \\
                             & test                     & 1,641           & 7,610   & 4,138          & 11,562  & 7.05     & 4.64     & 2,183           & 1.33     & 3.49     \\ \cmidrule{2-11}
                             & total                    & 22,765          & 177,192 & 49,870         & 163,206 & 7.18     & 5.79     & 30,926          & 1.35     & 4.29     \\ \midrule
\multirow{4}{*}{EmpathyCC}   & train                    & 17,879          & 122,597 & 35,294         & 73,748  & 4.12     & 6.86     & 19,234          & 1.08     & 6.37     \\
                             & valid                    & 2,347           & 7,631   & 4,125          & 9,720   & 4.14     & 3.25     & 2,540           & 1.08     & 3.00     \\
                             & test                     & 2,165           & 7,924   & 4,402          & 8,932   & 4.13     & 3.66     & 2,344           & 1.08     & 3.38     \\ \cmidrule{2-11}
                             & total                    & 22,391          & 138,152 & 43,821         & 92,400  & 4.13     & 4.59     & 24,118          & 1.08     & 4.25     \\ \midrule
\multirow{4}{*}{PersonaCC}   & train                    & 8,798           & 150,818 & 41,579         & 129,404 & 14.71    & 17.14    & 20,648          & 2.35     & 7.30     \\
                             & valid                    & 956             & 10,289  & 4,406          & 14,916  & 15.60    & 10.76    & 2,278           & 2.38     & 4.52     \\
                             & test                     & 933             & 10,163  & 4,407          & 14,474  & 15.51    & 10.89    & 2,195           & 2.35     & 4.63     \\ \cmidrule{2-11}
                             & total                    & 10,687          & 171,270 & 50,392         & 158,794 & 15.27    & 12.93    & 25,121          & 2.36     & 5.48     \\ \midrule
\multirow{4}{*}{KnowledgeCC} & train                    & 17,538          & 210,940 & 54,210         & 158,773 & 9.05     & 12.03    & 32,015          & 1.83     & 6.59     \\
                             & valid                    & 1,740           & 12,666  & 5,749          & 15,834  & 9.10     & 7.28     & 3,138           & 1.80     & 4.04     \\
                             & test                     & 1,694           & 12,213  & 5,915          & 15,358  & 9.07     & 7.21     & 3,023           & 1.78     & 4.04     \\ \cmidrule{2-11}
                             & total                    & 20,972          & 235,819 & 65,874         & 189,965 & 9.07     & 8.84     & 38,176          & 1.80     & 4.89     \\  \bottomrule                       
\end{tabular}

\end{adjustbox}
\caption{Statistics of sub-dataset of DialogCC. I./D. and I./S. denote images by dialogue and images by an image-sharing utterance, respectively. U./D. denotes utterances by a dialogue.}
\label{supp_tab:sub_dialogcc_stat}

\end{table*}}

\subsection{Full Statistics of DialogCC}

Table~\ref{supp_tab:full_dataset_stat} presents a comprehensive comparison of the statistics for DialogCC against existing datasets, namely PhotoChat, MMDD, and MMDialog. DialogCC is constructed from five source dialogue datasets: Persona-Chat, EmpatheticDialogues, Blended Skill Talk, DailyDialog, and Wizard-of-Wikipedia. As a result, DialogCC consists of five sub-datasets: BlendedCC, DailyCC, EmpathyCC, PersonaCC, and KnowledgeCC. Taking PersonaCC as an example, this dataset is formulated by aligning images from the CC3M collection with the Persona-Chat dataset, achieved using our proposed automatic pipeline. The statistics of five sub-datasets are presented in Table~\ref{supp_tab:sub_dialogcc_stat}.

\begin{figure*}[t]
    \centering
    \includegraphics[width=\textwidth]{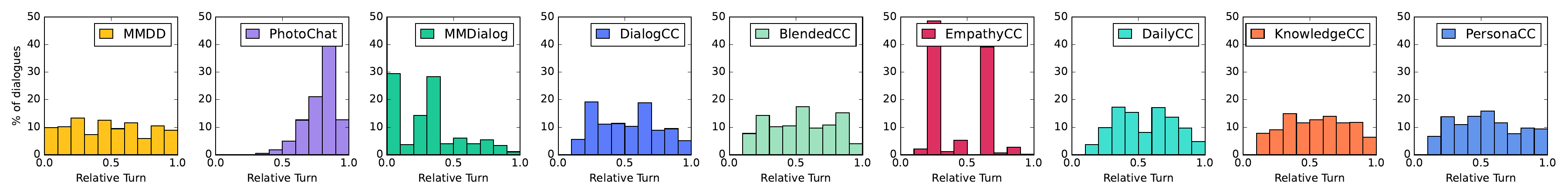}
    \caption{Comparison of DialogCC with other multi-modal dialogue datasets: PhotoChat~\cite{zang2021photochat}, MMDD~\cite{lee2021constructing}, and MMDialog~\cite{feng2022mmdialog}, in terms of the distribution of image-sharing moments. The x-axis and y-axis represent the relative turn ratio and \% of dialogues, respectively. We also show the distribution of a subset of DialogCC: \textcolor{blendedCC}{BlendedCC}, \textcolor{empathyCC}{EmpathyCC}, \textcolor{dailyCC}{DailyCC}, \textcolor{knowledgeCC}{KnowledgeCC}, and \textcolor{personaCC}{PersonaCC}.}
    \label{supp_fig:turn_dist}
\end{figure*}

\subsection{Image-Sharing Moment Distribution}

In Figure~\ref{supp_fig:turn_dist}, we analyze the distribution of turns at which image-sharing occurs across various datasets. Unlike PhotoChat and MMDialog, DialoCC demonstrates that the moments that images are shared are evenly distributed throughout the conversation turns. This suggests that models trained on our dataset may better understand the optimal moments for image-sharing across diverse dialogue turns.
Compared with MMDD, the turn distribution for image sharing in MMDD is also even. However, it's notable that in MMDD, images can be seen even in the initial dialogue turn. As highlighted in Section~\ref{supp_sec:comparison}, MMDD might not fully represent an image-sharing dataset, given it also encompasses image-grounded dialogues. This observation suggests that during the creation of the MMDD dataset, images were potentially matched with single utterances based on image-text similarity via the VSRN model. Such an approach might not truly reflect humans' cognitive processes when sharing images in real-life conversations.
In contrast, DialoCC leveraged GPT-4 to determine appropriate moments to share an image in specific dialogues. This method results in a more naturally flowing dialogue with greater turn relevance, as shown in Figure 5. Consequently, in DialoCC, we can affirmatively state that no images are shared during the initial turn of our dialogues, unlike MMDD.

{\renewcommand{\arraystretch}{1.35}
\begin{table}[t]
\centering
\begin{adjustbox}{width=\columnwidth}

\begin{tabular}{lccccccc}
\toprule
& \multicolumn{1}{l}{} & \multicolumn{3}{c}{Image Caption}                   & \multicolumn{3}{c}{Dialogue}                           \\ \cmidrule{3-8}
Dataset                    & Type           & \# hyp         & \# unigram      & \# bigram        & \# hyp         & \# unigram       & \# bigram          \\ \midrule
\multirow{4}{*}{PhotoChat} & train          & 293            & 4,203           & 10,772           & 1,203          & 18,252           & 179,904            \\
                           & valid          & 72             & 1,001           & 2,059            & 348            & 4,994            & 32,883             \\
                           & test           & 74             & 1,000           & 2,034            & 351            & 5,066            & 33,456             \\ \cmidrule{2-8}
                           & total          & 439            & 6,204           & 14,865           & 1,902          & 28,312           & 246,243            \\ \midrule
\multirow{4}{*}{MMDD}      & train          & 1,832          & 11,571          & 95,918           & 2,168          & 23,264           & 298,517            \\
                           & valid          & 462            & 2,080           & 7,539            & 968            & 10,207           & 88,762             \\
                           & test           & 463            & 2,337           & 8,867            & 1,033          & 11,055           & 96,891             \\ \cmidrule{2-8}
                           & total          & 2,757          & 15,988          & 112,324          & 4,169          & 44,526           & 484,170            \\ \midrule
\multirow{4}{*}{MMDialog}  & train          & -              & -               & -                & 9,271          & 772,044          & 8,582,862          \\
                           & valid          & -              & -               & -                & 2,239          & 49,443           & 340,221            \\
                           & test           & -              & -               & -                & 2,247          & 49,310           & 339,883            \\ \cmidrule{2-8}
                           & total          & -              & -               & -                & 13,757         & 870,797          & 9,262,966          \\ \midrule
\multirow{4}{*}{DialogCC}  & train          & 3,020          & 18,623          & 241,047          & 4,061          & 62,961           & 953,730            \\
                           & valid          & 1,469          & 9,485           & 58,320           & 1,802          & 22,096           & 219,545            \\
                           & test           & 1,493          & 9,725           & 59,529           & 1,819          & 21,873           & 216,436            \\ \cmidrule{2-8}
                           & \textbf{total} & \textbf{5,982} & \textbf{37,833} & \textbf{358,896} & \textbf{7,682} & \textbf{106,930} & \textbf{1,389,711} \\  \bottomrule                       
\end{tabular}

\end{adjustbox}
\caption{We count the number of unique hypernyms from WordNet~\cite{miller1995wordnet} and words in dialogues and image captions. We filter out a hypernym if it appears less than ten times in both dialogues and image captions. \# hyp, \# unigram, and \# bigram denote the number of hypernyms, the number of unique unigrams, and the number of unique bigrams.
respectively}
\label{supp_tab:full_diversity_stat}

\end{table}}

\subsection{Diversity}

In Table~\ref{supp_tab:full_diversity_stat}, we compare the diversity of datasets with the number of unique hypernyms from WordNet~\cite{miller1995wordnet} and words in dialogues and image captions. As WordNet covers nouns, verbs, adjectives, and adverbs, we only count nouns by filtering out the hypernyms appearing less than ten times.
Compared to PhotoChat and MMDD, DialogCC contains the largest number of unique hypernyms and unique words in both image captions and dialogues. 
Unfortunately, MMDialog does not include captions, so we cannot determine the number of unique hypernyms and unique words from that dataset.
However, MMDialog has more hypernyms and unique words, likely attributed to its larger volume of dialogues. It's worth noting that despite MMDialog having the most extensive scale, its quality is subpar, as depicted in Figure 5.

\subsection{Rationale Distribution}

To gain a better understanding of the generated rationales, we conduct an analysis of their verb-noun patterns. Table~\ref{supp_tab:rationale_distribution} shows the rationale distribution obtained from GPT-4. We parse the rationales using spaCy~\cite{honnibal2020spacy} and extract the root verb along with its first direct noun object.
Since we constrain a rationale to start with ``To'' in the prompt, we only consider rationales with a ``To verb noun'' structure during this analysis.
Out of a total of 106,063 generated rationales, 102,554 rationales follow this structure, whereas 3,509 rationales contain more complex clauses (e.g., \textit{To show how he spent his relaxing weekend.}).

In this analysis, we observe that the verb ``provide'' is used most frequently. This indicates that image sharing is intended to provide additional information related to the context of the dialogue. The tendency to provide additional information through image sharing is also evident in the verbs ``show'' and ``share''.

\begin{figure}[t]
    \centering
    \includegraphics[width=\columnwidth]{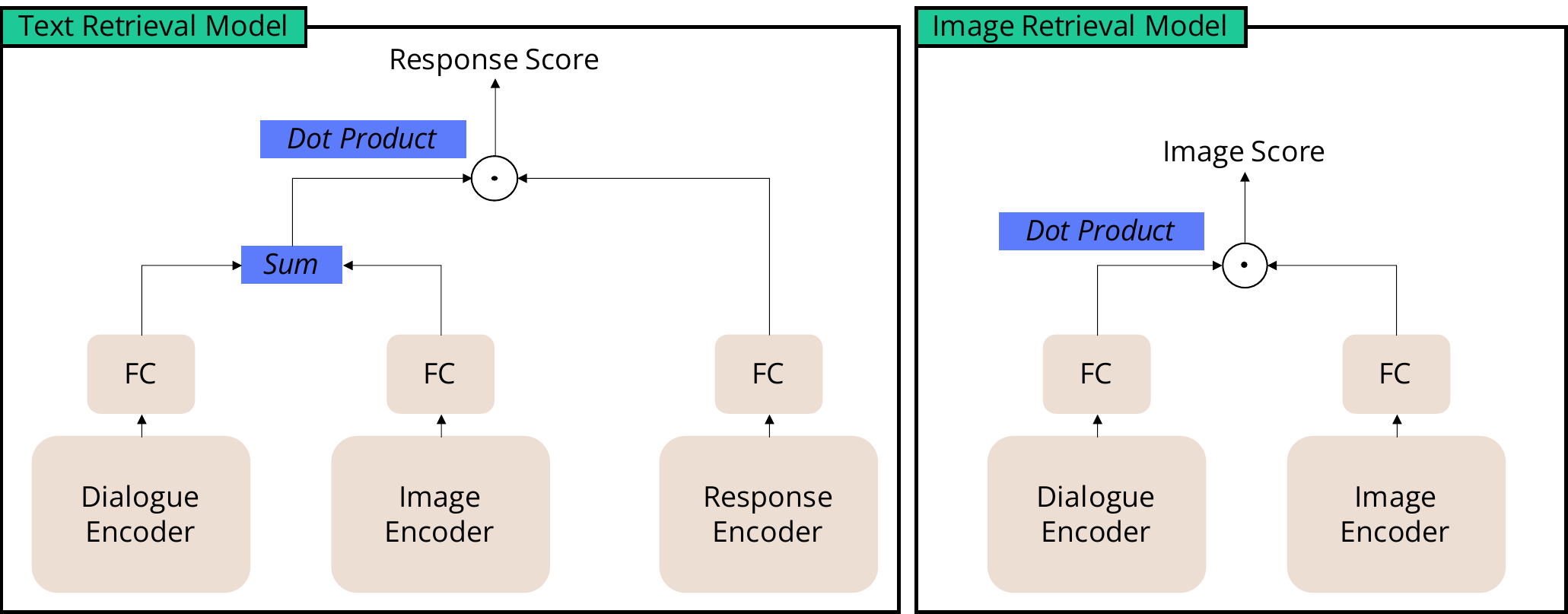}
    \caption{Architectures of two baseline models: Text retrieval and Image retrieval.}
    \label{supp_fig:basline}
\end{figure}

\subsection{More Examples of DialogCC} \label{supp_sec:more_examples}

We present more examples of DialogCC in Figure~\ref{supp_fig:case1}, Figure~\ref{supp_fig:case2}, Figure~\ref{supp_fig:case3}, Figure~\ref{supp_fig:case4}, and Figure~\ref{supp_fig:case5}.

\section{Details of Experimental Settings} \label{supp_sec:expr_setting}

\subsection{Baseline Models} \label{supp_sec:baselines}

As illustrated in Figure~\ref{supp_fig:basline}, we present the architecture of baseline models, which is the text retrieval model and image retrieval model. We provide a detailed description of baseline models below.

\paragraph{Text Retrieval Model.} 
The text retrieval model comprises three main components: the dialogue encoder, the response encoder, and the image encoder.
The dialogue encoder processes the entire dialogue history and transforms it into a fixed-size representation. To achieve this, we use the BERT model~\cite{devlin2018bert}. The dialogue history consists of up to three turns preceding the current turn. Each turn is concatenated using the \texttt{[SEP]} special token.
The response encoder is responsible for converting the response into a fixed-size representation. While it also utilizes the BERT model, the specific BERT version used here is different from that employed in the dialogue encoder. For both the dialogue and response encoders, after processing the text with BERT, we apply mean pooling to the text representations. The pooled representations are subsequently passed through a linear projection layer, which is then followed by the ReLU activation function~\cite{nair2010rectified}.
The image encoder is to extract feature vectors from images, and for this purpose, we utilize the CLIP-base model~\cite{radford2021learning}.
Once the feature vectors are extracted from the dialogue and images, we perform an element-wise addition of the image vectors and dialogue vectors. To compute the loss, we calculate the dot product between the response feature vector and the resulting summed vector.

\paragraph{Image Retrieval Model.}

The image retrieval model is composed of two main components: the dialogue encoder and the image encoder.
The dialogue encoder utilizes the BERT-base model to transform the dialogue into a representation. After encoding, we apply mean pooling to the text representations derived from this dialogue encoder.
For image representation, we employ the CLIP-base model.
Following the encoding processes, both the image and dialogue vectors are passed through separate linear projection layers, each followed by a ReLU activation function. To determine the loss, we calculate the dot product between the image feature vector and the dialogue vector.

\subsection{Implementation Details}

We implement baseline models based on PyTorch Lightning. All experiments are conducted on two A100 GPUs (40GB). To accelerate the training time, we apply distributed training to baselines. We follow the hyperparameter settings similar to the previous works~\cite{lee2021constructing,zang2021photochat}, which are described as follows:

\paragraph{Text retrieval.} In our experiment, we set the batch size to 256, the learning rate to 5e-5, and the gradient clipping value to 2.0. We use the AdamW optimizer with a cosine learning rate scheduler. We set the warm-up ratio as 0.1\% and weight decay as 0.2.

\paragraph{Image retrieval.} We set the batch size to 256. We also use the AdamW optimizer with an initial learning rate of 2e-5 and decaying 0.1\% at every 1,000 steps. We set the warm-up ratio as 0.1\%. 

\paragraph{Training.} Since our dataset contains several images per utterance, we randomly choose one image in each batch. We do not update the parameter of the image encoder.

\section{Further Experiments} \label{supp_sec:further_expr}

\subsection{Full Results} \label{supp_sec:full_rc_results}

Table~\ref{supp_tab:full_rec_result} shows the full results of model trained on PhotoChat~\cite{zang2021photochat}, MMDD~\cite{lee2021constructing}, MMDialog~\cite{feng2022mmdialog}, and DialogCC (ours). Then, we evaluate each trained model on four different datasets. We measure the average contributed performance only considering the out-of-domain datasets. For example, if the model is trained on the MMDD dataset, then we calculate the averaged contributed performance by evaluating this model on PhotoChat, MMDialog, and DialogCC.

\subsection{Breakdown Results in DialogCC}
We show the additional experiments in Table~\ref{supp_tab:sub_sub_ir} and Table~\ref{supp_tab:sub_sub_nrp}.
We evaluate the trained retrieval model on the sub-dataset of DialogCC to other sub-dataset of DialogCC.

\section{Human Evaluation Questionnaire} \label{supp_sec:human_eval_question}

This section presents the list of questions and multiple-choice options used for two human evaluations reported in Section 3.4: human ratings and head-to-head comparison. 

\begin{figure}[t]
    \centering
    \includegraphics[width=0.9\columnwidth]{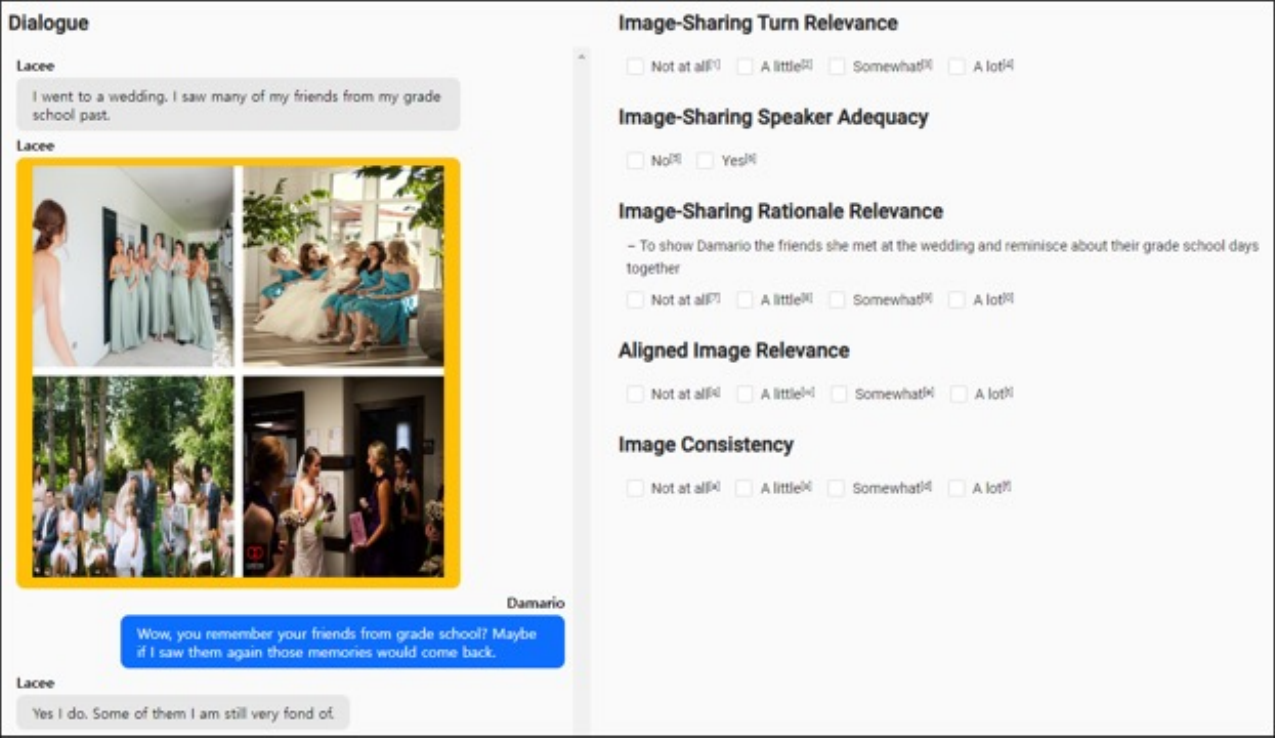}
    \caption{A screenshot of the human evaluation system for the human ratings.}
    \label{supp_fig:rating}
\end{figure}

\begin{figure}[t]
    \centering
    \includegraphics[width=0.9\columnwidth]{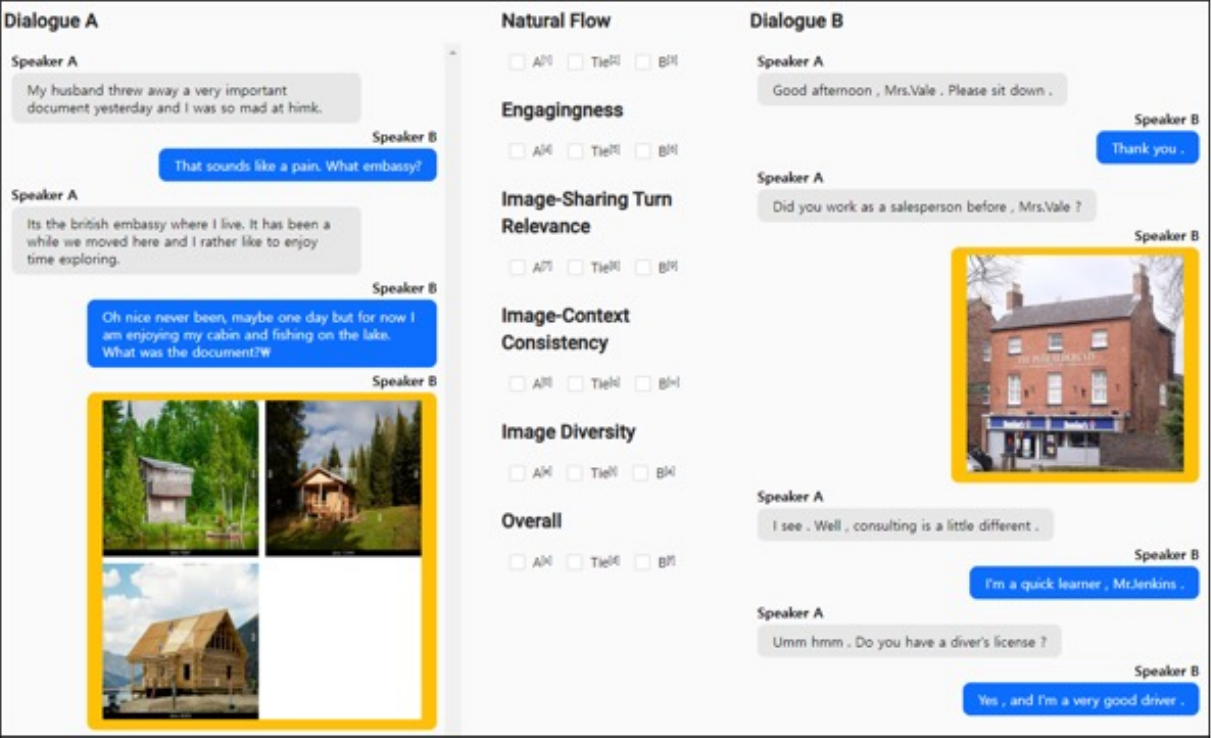}
    \caption{A screenshot of the human evaluation system for the head-to-head comparison.}
    \label{supp_fig:ab}
\end{figure}

{\renewcommand{\arraystretch}{1.35}
\begin{table}[t]
\centering
\begin{adjustbox}{width=\columnwidth}
\begin{tabular}{lcccccc} \toprule
                             &       & \makecell{Turn \\ Relevance} & \makecell{Rationale \\ Relevance} & \makecell{Aligned \\ Image Relevance} & \makecell{Image \\ Consistency} & \makecell{Speaker \\ (\% of Yes)} \\ \midrule
\multirow{2}{*}{DialogCC}    & Avg.  & 3.68           & 3.41      & 3.30          & 3.57              & 95.07          \\
                             & $\alpha$ & \cellcolor{teal!10}0.14           & \cellcolor{teal!25}0.39      & \cellcolor{teal!35}0.54          & \cellcolor{teal!35}0.50              & -              \\ \midrule
\multirow{2}{*}{KnowledgeCC} & Avg.  & 3.61           & 3.15      & 3.05          & 3.38              & 99.33          \\
                             & $\alpha$ & \cellcolor{teal!10}0.14           & \cellcolor{teal!25}0.38      & \cellcolor{teal!45}0.64          & \cellcolor{teal!35}0.59              & -              \\ \midrule
\multirow{2}{*}{PersonaCC}   & Avg.  & 3.84           & 3.71      & 3.69          & 3.80              & 92.67          \\
                             & $\alpha$ & \cellcolor{teal!10}-0.03          & \cellcolor{teal!25}0.24      & \cellcolor{teal!25}0.27          & \cellcolor{teal!35}0.59              & -              \\ \midrule
\multirow{2}{*}{EmpathyCC}   & Avg.  & 3.71           & 3.43      & 3.30          & 3.59              & 97.33          \\
                             & $\alpha$ & \cellcolor{teal!10}0.12           & \cellcolor{teal!25}0.31      & \cellcolor{teal!35}0.55          & \cellcolor{teal!35}0.56              & -              \\ \midrule
\multirow{2}{*}{BlendedCC}   & Avg.  & 3.67           & 3.49      & 3.37          & 3.61              & 88.67          \\
                             & $\alpha$ & \cellcolor{teal!10}0.11           & \cellcolor{teal!35}0.45      & \cellcolor{teal!25}0.32          & \cellcolor{teal!10}0.16              & -              \\ \midrule
\multirow{2}{*}{DailyCC}     & Avg.  & 3.54           & 3.27      & 3.11          & 3.47              & 97.33          \\
                             & $\alpha$ & \cellcolor{teal!10}0.16           & \cellcolor{teal!25}0.36      & \cellcolor{teal!35}0.58          & \cellcolor{teal!35}0.45              & -             
\\  \bottomrule                       
\end{tabular}

\end{adjustbox}
\caption{Breakdown human evaluation results.}
\label{supp_tab:full_human_eval}

\end{table}}

{\renewcommand{\arraystretch}{1.35}
\begin{table*}[t]
\centering
\begin{adjustbox}{width=\textwidth}

\begin{tabular}{@{}lcccccccccccccccc@{}}
\toprule
Eval          & \multicolumn{4}{c}{MMDD}                                          & \multicolumn{4}{c}{PhotoChat}                                     & \multicolumn{4}{c}{MMDialog}                                      & \multicolumn{4}{c}{DialogCC}                                      \\ \cmidrule(l){2-17} 
Train Dataset & R@1            & R@5            & R@10           & MRR            & R@1            & R@5            & R@10           & MRR            & R@1            & R@5            & R@10           & MRR            & R@1            & R@5            & R@10           & MRR            \\ \midrule
\multicolumn{17}{l}{\bluecell \textit{Image Retrieval}}                                                                                                                                                                                                                                                          \\
MMDD          & 4.20           & 14.26          & 22.22          & 10.91          & 7.90           & 24.01          & 35.34          & 17.16          & 5.06           & 17.47          & 27.45          & 12.91          & 6.72           & 23.38          & 36.07          & 16.35          \\
PhotoChat     & 3.13           & 9.52           & 16.85          & 8.58           & 5.41           & 23.70          & 38.67          & 15.87          & 3.43           & 13.10          & 22.07          & 10.14          & 4.36           & 16.45          & 27.21          & 12.23          \\
MMDialog      & 3.47           & 13.14          & 20.36          & 10.01          & 7.17           & 23.91          & 38.05          & 16.79          & \textbf{19.79} & \textbf{51.13} & \textbf{66.56} & \textbf{34.66} & 8.91           & 29.70          & 43.68          & 20.06          \\
DialogCC      & \textbf{6.45}  & \textbf{17.33} & \textbf{26.27} & \textbf{13.32} & \textbf{13.51} & \textbf{37.32} & \textbf{51.14} & \textbf{25.63} & 10.97          & 32.53          & 45.66          & 22.40          & \textbf{17.09} & \textbf{46.53} & \textbf{62.29} & \textbf{31.36} \\
\multicolumn{17}{l}{\bluecell \textit{Next Response Prediction}}                                                                                                                                                                                                                                                 \\
MMDD          & \textbf{19.97} & \textbf{40.63} & \textbf{50.93} & \textbf{30.40} & 7.88           & 21.25          & 29.45          & 15.91          & 8.33           & 24.08          & 36.14          & 17.62          & 16.44          & 41.74          & 55.16          & 29.16          \\
PhotoChat     & 6.40           & 19.09          & 31.69          & 14.49          & \textbf{9.39}  & \textbf{25.03} & \textbf{39.05} & \textbf{19.08} & 5.18           & 17.81          & 28.57          & 13.18          & 7.59           & 24.34          & 36.53          & 17.22          \\
MMDialog      & 9.67           & 27.10          & 39.01          & 19.67          & 8.95           & 24.70          & 34.41          & 17.92          & \textbf{34.21} & \textbf{61.22} & \textbf{72.88} & \textbf{46.98} & 17.43          & 40.10          & 52.51          & 29.01          \\
DialogCC      & 18.46          & 32.52          & 42.09          & 26.54          & 8.09           & 20.50          & 29.88          & 16.26          & 12.69          & 30.16          & 42.02          & 22.72          & \textbf{40.64} & \textbf{71.46} & \textbf{81.99} & \textbf{54.61} \\ \bottomrule
                   
\end{tabular}

\end{adjustbox}
\caption{We report the full results on the next response prediction and image retrieval tasks. The model with the best performance is indicated in \textbf{bold}.}
\label{supp_tab:full_rec_result}

\end{table*}}

{\renewcommand{\arraystretch}{1.35}
\begin{table*}[!t]
\centering
\begin{adjustbox}{width=\textwidth}

\begin{tabular}{lcccccccccccccccccccc}
\toprule
Eval $\rightarrow$       & \multicolumn{4}{c}{BlendedCC}                                     & \multicolumn{4}{c}{DailyCC}                                       & \multicolumn{4}{c}{EmpathyCC}                                     & \multicolumn{4}{c}{PersonaCC}                                     & \multicolumn{4}{c}{KnowledgeCC}                                   \\ \cmidrule{2-21}
Train $\downarrow$      & R@1            & R@5            & R@10           & MRR            & R@1            & R@5            & R@10           & MRR            & R@1            & R@5            & R@10           & MRR            & R@1            & R@5            & R@10           & MRR            & R@1            & R@5            & R@10           & MRR            \\ \midrule
BlendedCC   & 16.60          & 47.67          & 65.02          & 31.31          & 12.28          & 34.16          & 49.20          & 23.98          & 12.87          & 38.31          & 54.65          & 25.79          & 12.69          & 36.26          & 53.21          & 25.10          & \underline{19.06}    & \underline{49.43}    & \underline{66.51}    & \underline{33.68}    \\
DailyCC     & 16.19          & 43.48          & 60.43          & 29.77          & \textbf{21.22} & \textbf{52.76} & \textbf{68.86} & \textbf{36.25} & \underline{17.68}    & \underline{46.46}    & \underline{62.05}    & \underline{31.52}    & 10.40          & 31.17          & 47.19          & 21.76          & 15.90          & 46.11          & 63.61          & 30.58          \\
EmpathyCC   & \underline{19.89}    & 45.89          & 62.35          & \underline{32.88}    & 13.92          & \underline{42.08}    & \underline{60.45}    & \underline{28.09}    & \textbf{19.80} & \textbf{51.22} & \textbf{67.64} & \textbf{34.82} & 11.63          & 34.15          & 51.10          & 23.54          & 16.00          & 45.38          & 62.05          & 30.32          \\
PersonaCC   & 17.56          & \underline{49.11}    & \underline{66.46}    & 32.66          & 12.19          & 32.30          & 47.33          & 22.99          & 13.62          & 38.90          & 54.06          & 26.40          & \underline{14.18}    & \underline{39.86}    & \textbf{57.42} & \underline{27.29}    & 17.72          & 47.80          & 65.30          & 32.21          \\
KnowledgeCC & \textbf{22.91} & \textbf{54.46} & \textbf{69.75} & \textbf{37.38} & \underline{15.61}    & 39.55          & 53.60          & 27.83          & 14.96          & 39.06          & 53.03          & 26.99          & \textbf{14.75} & \textbf{41.22} & \underline{57.02}    & \textbf{28.03} & \textbf{26.83} & \textbf{65.14} & \textbf{79.45} & \textbf{43.63}
  \\  \bottomrule                       
\end{tabular}

\end{adjustbox}
\caption{We report the image retrieval performance on BlendedCC, DailyCC, EmpathyCC, PersonaCC, and KnowledgeCC. The model with the best performance is indicated in \textbf{bold}, while the second best is \underline{underlined}.}
\label{supp_tab:sub_sub_ir}

\end{table*}}

{\renewcommand{\arraystretch}{1.35}
\begin{table*}[t]
\centering
\begin{adjustbox}{width=\textwidth}

\begin{tabular}{lcccccccccccccccccccc}
\toprule
Eval $\rightarrow$       & \multicolumn{4}{c}{BlendedCC}                                     & \multicolumn{4}{c}{DailyCC}                                       & \multicolumn{4}{c}{EmpathyCC}                                     & \multicolumn{4}{c}{PersonaCC}                                     & \multicolumn{4}{c}{KnowledgeCC}                                   \\ \cmidrule{2-21}
Train $\downarrow$      & R@1            & R@5            & R@10           & MRR            & R@1            & R@5            & R@10           & MRR            & R@1            & R@5            & R@10           & MRR            & R@1            & R@5            & R@10           & MRR            & R@1            & R@5            & R@10           & MRR            \\ \midrule
BlendedCC   & \textbf{16.51} & \underline{44.15}    & \textbf{58.94} & \textbf{30.19} & 8.27           & 25.00          & 37.43          & 17.82          & 8.20           & 23.75          & 37.52          & 17.67          & \underline{10.00}    & \underline{28.66}    & \underline{41.08}    & \underline{20.08}    & \underline{17.18}    & \underline{45.77}    & \underline{62.78}    & \underline{31.29}    \\
DailyCC     & 10.98          & 33.60          & 47.16          & 22.71          & \textbf{23.70} & \textbf{55.71} & \textbf{68.69} & \textbf{38.57} & 8.64           & \underline{27.99}    & \underline{42.33}    & 19.55          & 6.30           & 21.41          & 32.63          & 15.17          & 13.97          & 42.33          & 58.49          & 27.68          \\
EmpathyCC   & 12.06          & 35.39          & 51.69          & 24.50          & \underline{9.62}     & \underline{27.20}    & \underline{39.63}    & \underline{19.62}    & \textbf{18.78} & \textbf{47.82} & \textbf{62.68} & \textbf{32.79} & 7.62           & 23.64          & 35.83          & 16.80          & 13.16          & 39.50          & 56.51          & 26.47          \\
PersonaCC   & 14.50          & 41.13          & 54.06          & 27.28          & 7.16           & 20.99          & 31.41          & 15.68          & 6.14           & 19.95          & 30.78          & 14.45          & \textbf{10.32} & \textbf{30.17} & \textbf{43.36} & \textbf{21.28} & 12.07          & 37.66          & 53.92          & 24.78          \\
KnowledgeCC & \underline{16.37}    & \textbf{44.44} & \underline{57.07}    & \underline{30.00}    & 9.47           & 26.25          & 38.43          & 19.16          & \underline{9.73}     & 27.79          & 40.39          & \underline{19.83}    & 8.90           & 26.11          & 37.15          & 18.34          & \textbf{25.19} & \textbf{62.68} & \textbf{75.97} & \textbf{41.79}

  \\  \bottomrule                       
\end{tabular}

\end{adjustbox}
\caption{We report the next response prediction performance on BlendedCC, DailyCC, EmpathyCC, PersonaCC, and KnowledgeCC. The model with the best performance is indicated in \textbf{bold}, while the second best is \underline{underlined}.}
\label{supp_tab:sub_sub_nrp}

\end{table*}}

\subsection{Human Ratings} \label{supp_sec:human_rating}

\begin{itemize}
    \item \textbf{Image-Sharing Turn Relevance:} Do you think the image-sharing turn in the given dialogue is appropriate?
    \begin{description}
        \item [Options:] 1: Not at all / 2: A little / 3: Somewhat / 4: A lot
    \end{description}
    
    \item \textbf{Image-Sharing Speaker Adequacy:} Do you think the speaker who shared the image in the given dialogue is appropriate? 
    \begin{description}
        \item [Options:] No / Yes
    \end{description}
    
    \item \textbf{Image-Sharing Rationale Relevance:} Do you think the reason for sharing the image in the given dialogue is valid? 
    \begin{description}
        \item [Options:] 1: Not at all / 2: A little / 3: Somewhat / 4: A lot
    \end{description}
    
    \item \textbf{Aligned Image Relevance:} How relevant do you think the aligned images are based on the dialogue context? 
    \begin{description}
        \item [Options:] 1: Not at all / 2: A little / 3: Somewhat / 4: A lot
    \end{description}
    
    \item \textbf{Image Consistency:} How consistent do you think there is between aligned images?
    \begin{description}
        \item [Options:] 1: Not at all / 2: A little / 3: Somewhat / 4: A lot
    \end{description}
    
\end{itemize}

\subsection{Head-to-Head Comparison} \label{supp_sec:head_to_head}

\begin{itemize}
    \item \textbf{Natural Flow:} Which dialogue has a more natural flow?
    \begin{description}
        \item [Options:] A / Tie / B
    \end{description}
    
    \item \textbf{Engagingness:} Which dialogue has more interesting and engaging?
    \begin{description}
        \item [Options:] A / Tie / B
    \end{description}
    
    \item \textbf{Image-Sharing Turn Relevance:} Which dialogue has a more appropriate image-sharing turn?
    \begin{description}
        \item [Options:] A / Tie / B
    \end{description}
    
    \item \textbf{Image-Dialogue Consistency:} Which dialogue is more consistent between aligned images and dialogue context? 
    \begin{description}
        \item [Options:] A / Tie / B
    \end{description}
    
    \item \textbf{Image Diversity:} Which dialogue has more diverse images?
    \begin{description}
        \item [Options:] A / Tie / B
    \end{description}

    \item \textbf{Overall:} Which dialogue has higher quality overall?
    \begin{description}
        \item [Options:] A / Tie / B
    \end{description}
    
\end{itemize}

\section{Human Evaluation System}

We show a screenshot of the human evaluation system in Figure~\ref{supp_fig:rating} and Figure~\ref{supp_fig:ab}. We implement this system using Label Studio~\cite{LabelStudio}.

\section{Details of Human Evaluation} \label{supp_sec:detail_human_eval}
We recruited three individuals, unknown to us, who are either graduate or undergraduate students. Prior to participating in the experiment, they were provided with comprehensive instruction on the task, an overview of the multi-modal dialogue dataset, and a detailed explanation of the evaluation criteria. This preparatory phase lasted approximately one hour. The detailed results of the human evaluation are presented in Table~\ref{supp_tab:full_human_eval}.

\begin{figure}[!t]
    \centering
    \includegraphics[width=\columnwidth]{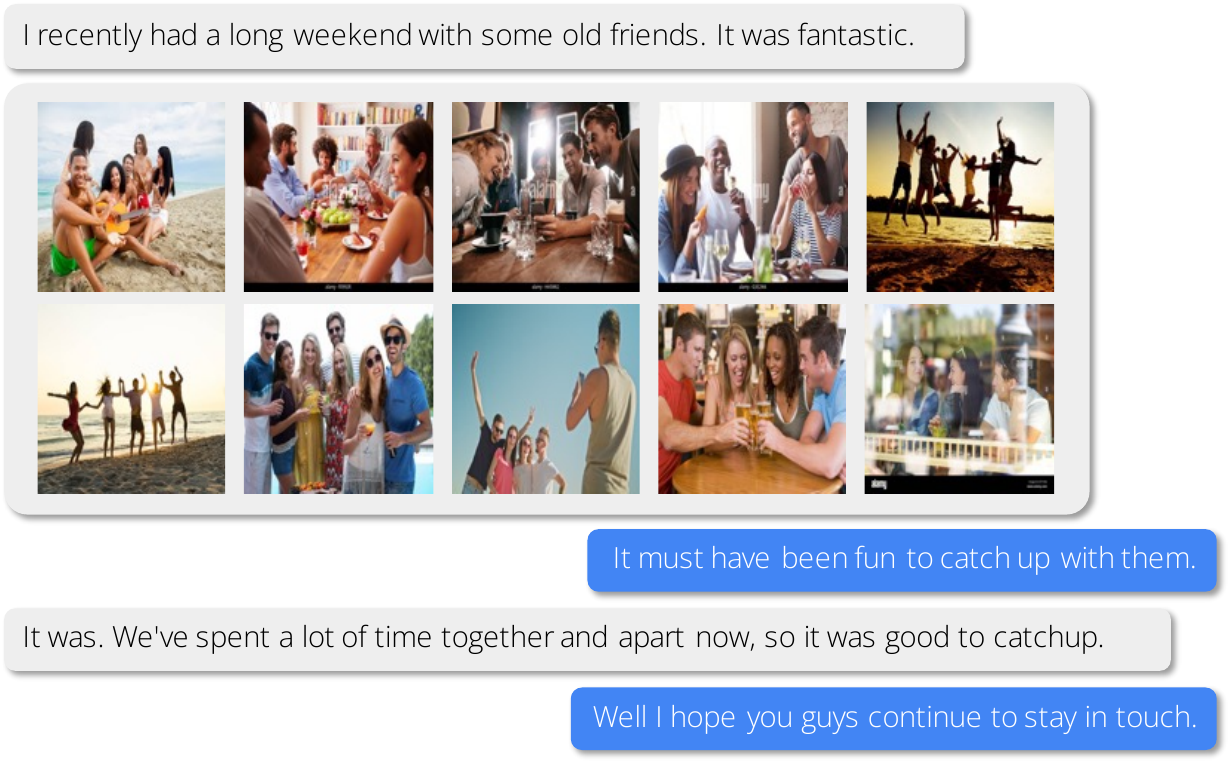}
    \caption{Case 1: An example of DialogCC.}
    \label{supp_fig:case1}
\end{figure}

\begin{figure}[!t]
    \centering
    \includegraphics[width=\columnwidth]{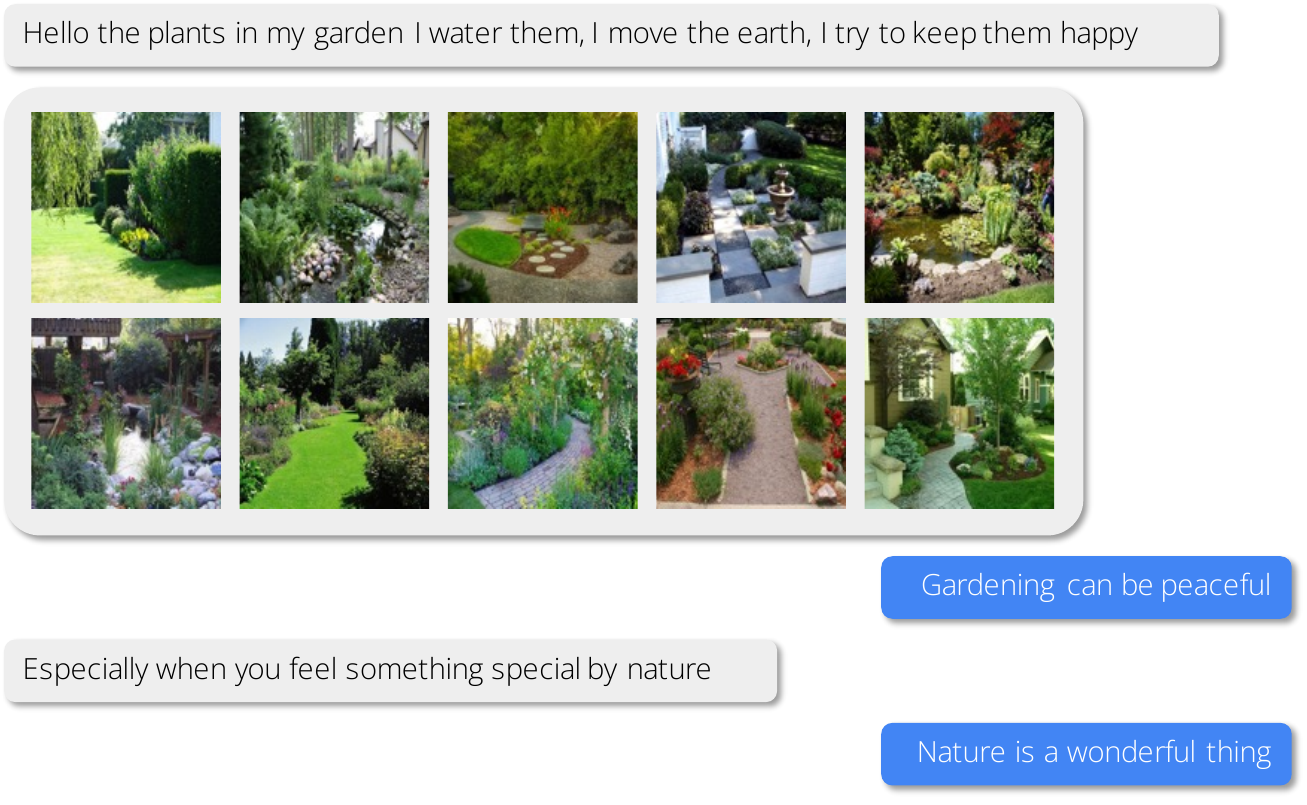}
    \caption{Case 2: An example of DialogCC.}
    \label{supp_fig:case2}
\end{figure}

\begin{figure}[!t]
    \centering
    \includegraphics[width=\columnwidth]{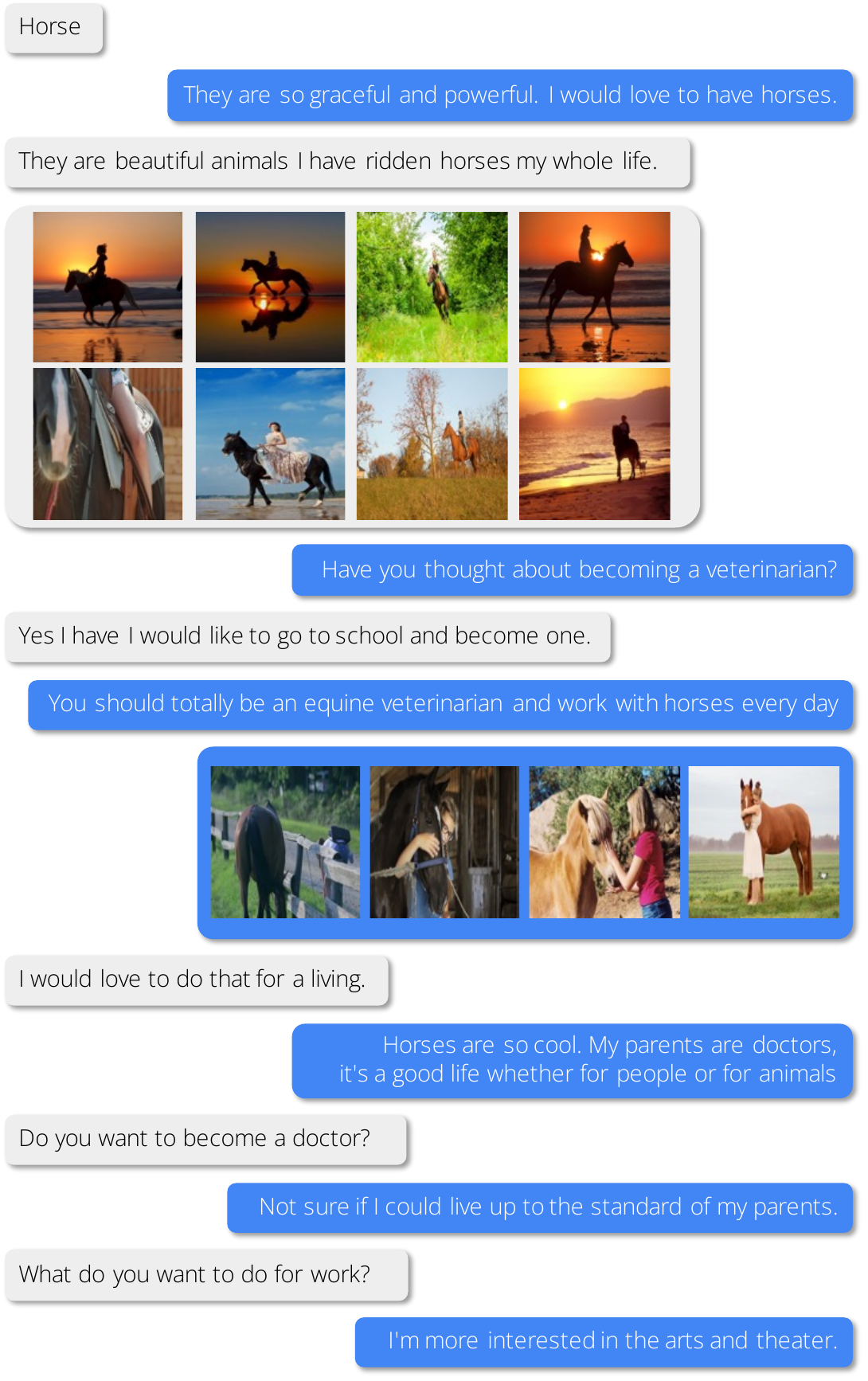}
    \caption{Case 3: An example of DialogCC.}
    \label{supp_fig:case3}
\end{figure}

\begin{figure}[!t]
    \centering
    \includegraphics[width=\columnwidth]{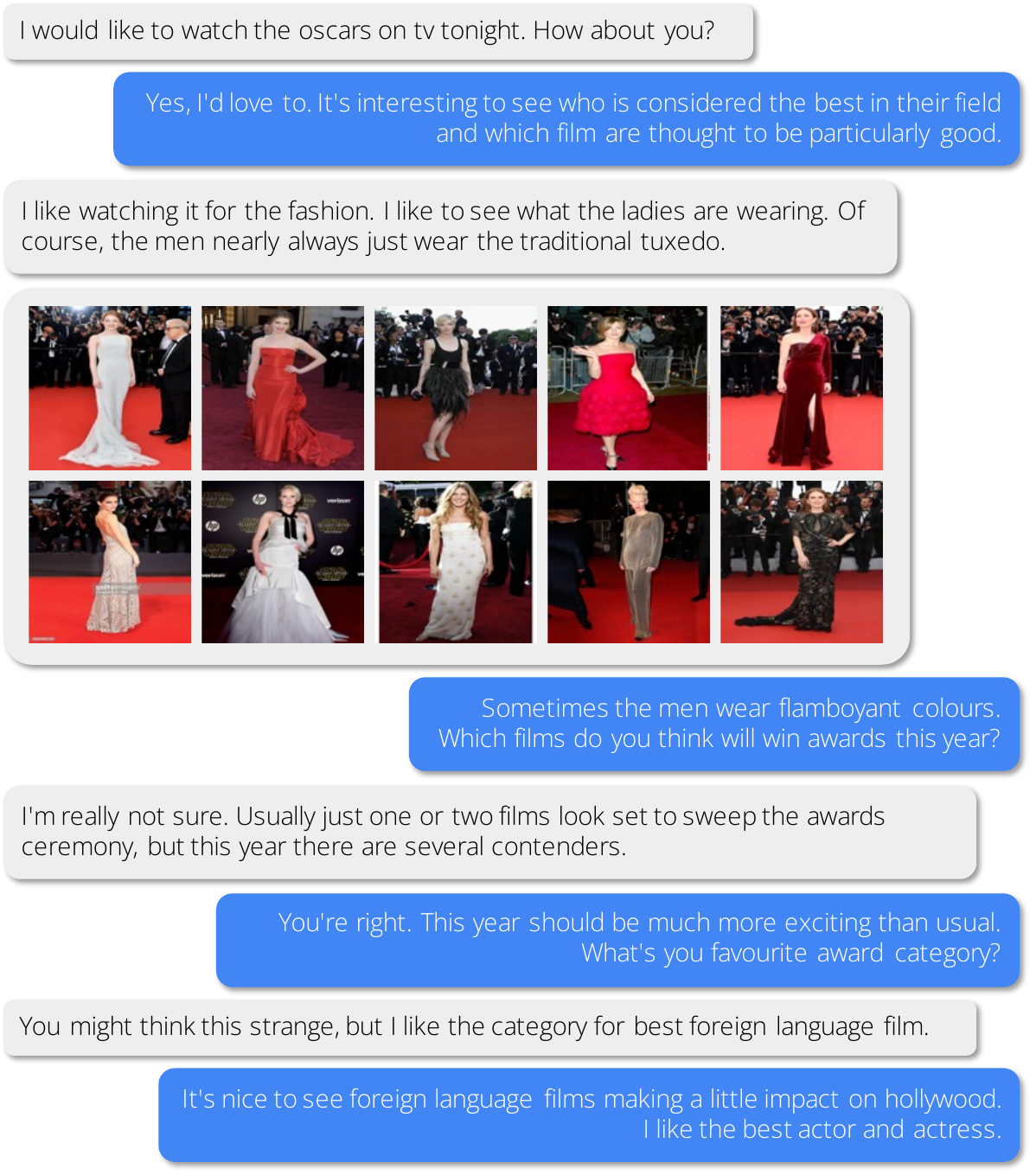}
    \caption{Case 4: An example of DialogCC.}
    \label{supp_fig:case4}
\end{figure}

\begin{figure}[!t]
    \centering
    \includegraphics[width=\columnwidth]{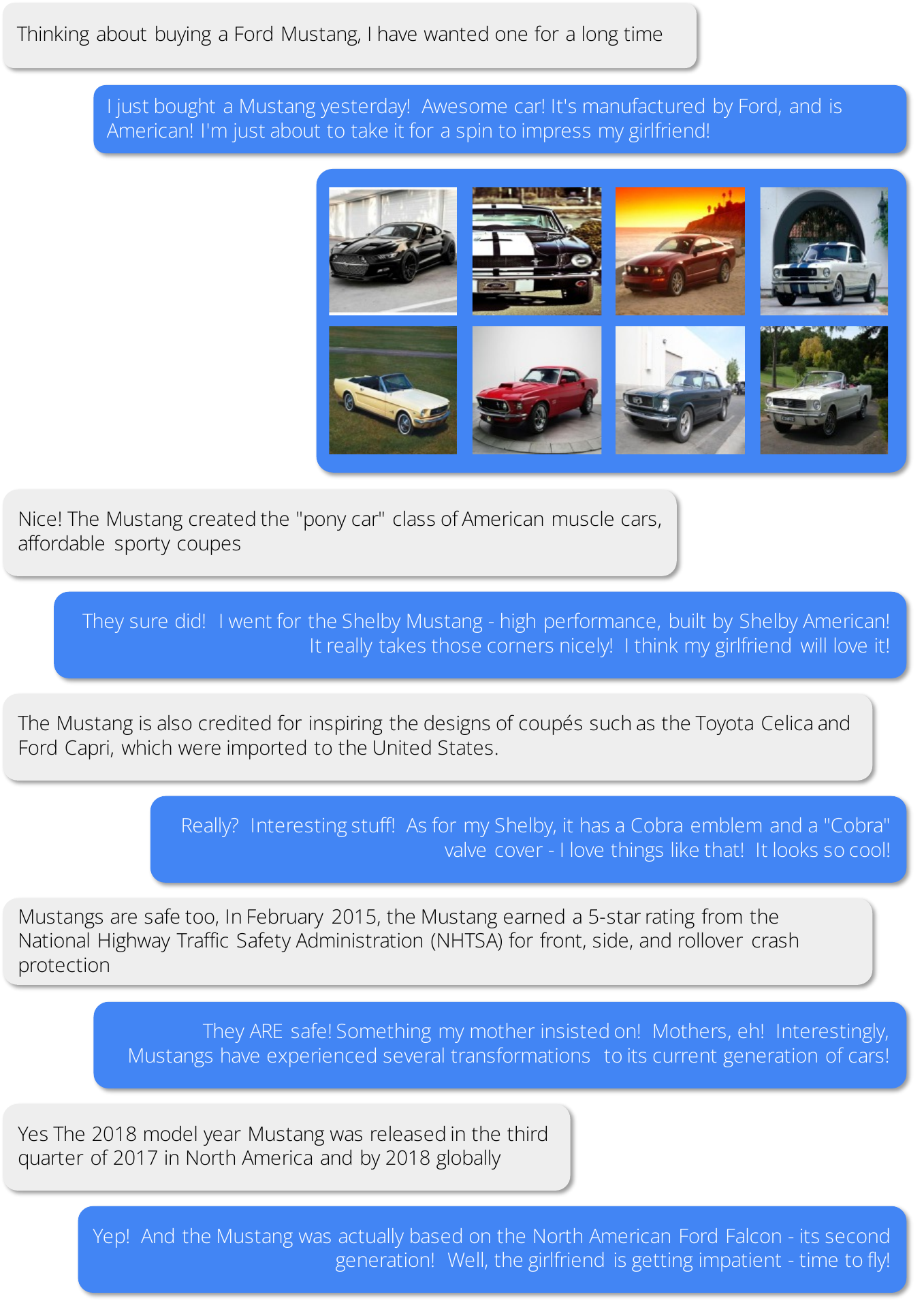}
    \caption{Case 5: An example of DialogCC.}
    \label{supp_fig:case5}
\end{figure}

\begin{figure*}[t]
    \centering
    \includegraphics[width=\textwidth]{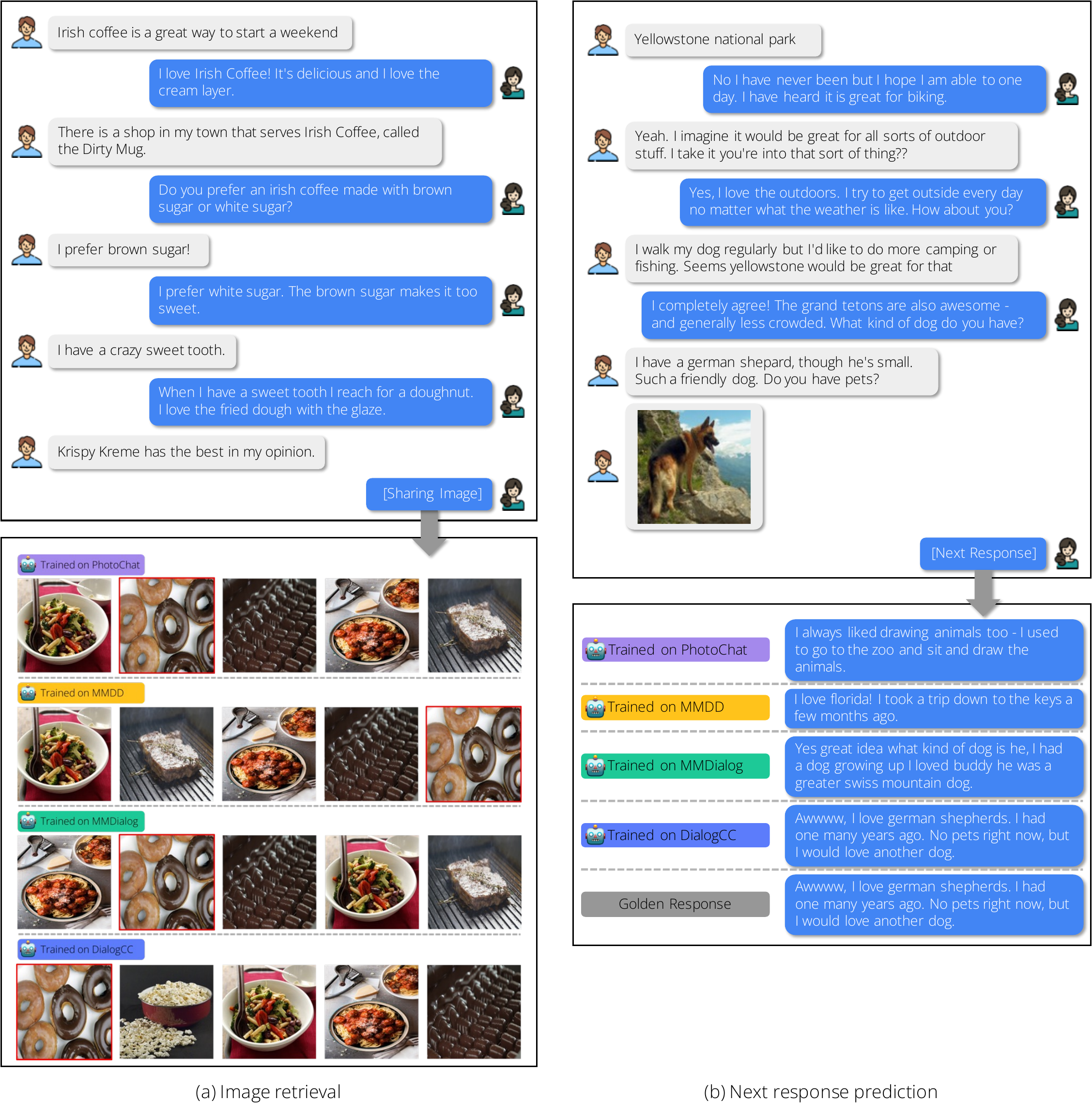}
    \caption{Two examples of retrieved results (\ie (a) image retrieval and (b) next response prediction) from four different models. Each provided dialogue is from the DialogCC dataset. In (a), we display the top-5 ranked images from left to right, with the ground-truth image marked in red. In (b), only the top-1 ranked next response is shown. Note that neither the \texttt{[Sharing Image]} turn nor the \texttt{[Next Response]} turn is provided to the model's input during the inference stage.}
    \label{supp_fig:case_study_dialogcc}
\end{figure*}

\begin{table*}[t]
\centering
\begin{adjustbox}{width=0.9\textwidth}
\begin{tabular}{@{}llrl@{}}
\toprule
Verb & Object & Count & Example \\
\midrule
provide & representation &  13,181 &               To provide a visual representation of yoga practice and the use of a yoga mat \\
        & evidence &   7,108 &                  To provide evidence of the value his company adds and support his argument \\
        & example &   3,387 &                  To provide a visual example of the kids' behavior that led to her yelling. \\
        & context &   1,802 &                       To provide context and show the positive change in the city's policy. \\
\midrule
show & example &   3,220 &                          To show an example of a craft Delfina made using Dollar Tree items \\
        & excitement &    761 &                             To show her excitement and happiness about having a little girl \\
        & type &    619 &                               To show the type of tins used for making cupcakes in the past \\
        & appreciation &    607 &             To show appreciation for his friends and emphasize their importance in his life \\
\midrule
share & image &   1,350 &                      To share a beautiful image of Hawaii that she remembers from her trip. \\
        & experience &   1,241 &                         To share a personal experience and highlight the beauty of Savannah \\
        & memory &    868 &                             To share a memory of their wedding day or a picture of his wife \\
        & picture &    751 &                               To share a picture of the delicious pasta Denese's wife makes \\
\midrule
give & idea &   1,081 &                       To give an idea of her living situation and the cost of her apartment \\
        & representation &    546 &                 To give Tyana a visual representation of the history of horse domestication \\
        & example &    212 &                          To give an example of the Beard of the Year award and its winners. \\
        & visual &    195 &                           To give Brenner a visual of the Acura to compare with the Integra \\
\midrule
showcase & skills &    518 &                            To showcase Tre's dancing skills or the type of dance they enjoy \\
        & variety &    235 &                        To showcase the variety of species Courtlyn keeps in their aquariums \\
        & work &    182 &                  To showcase his work and give Mary a better understanding of what he does. \\
        & passion &    170 &                         To showcase her passion for dancing and her favorite Disney moment. \\
\midrule
illustrate & concept &    251 &                   To illustrate the concept of hydraulic hybrids and how they store energy. \\
        & difference &    199 &             To illustrate the difference in the mountain scenery between January and April. \\
        & process &    107 &                        To illustrate the batting process and the pitcher's role in the game \\
        & connection &     94 &                To illustrate the connection between the company's name and its inspiration. \\
\midrule
emphasize & importance &    211 &                  To emphasize the importance of high-quality ingredients in Italian cooking \\
        & love &     50 &                         To emphasize her love for 2pac and how it complements her black car \\
        & popularity &     42 &                               To emphasize the popularity of My Little Pony toys in the 80s \\
        & preference &     38 &                             To emphasize the preference for a kitten as a pet over a snake. \\
\midrule
support & statement &    127 &                    To support their statement about liking pop music and finding it lovely. \\
        & argument &     30 &  To support the argument about the lack of educational programs and poorly done news shows. \\
        & claim &     27 &                               To support his claim and provide evidence for his prediction. \\
        & opinion &     26 &                     To support his opinion about Professor Wood and provide visual evidence \\
\midrule
express & interest &     45 &              To express his interest in trying mountain biking as another alternative sport \\
        & love &     32 &                                     To express her love for McDonald's breakfast and coffee \\
        & gratitude &     30 &                   To express gratitude and acknowledge the teacher's role in their success. \\
        & excitement &     23 &                           To express her excitement and share the news of winning the prize \\
\midrule
demonstrate & process &     43 &                      To demonstrate the process of adding a web page to the favorites list. \\
        & skills &     38 &                              To demonstrate her juggling skills and her work in the circus. \\
        & technique &     23 &                            To demonstrate the technique of playing the guitar in rock music \\
        & ability &     21 &                             To demonstrate the cat's ability to see in low light conditions \\
\midrule
confirm & order &     31 &                                  To confirm the order and show the specific items requested \\
        & details &     29 &             To confirm the booking details and provide a visual summary of the reservation. \\
        & time &     23 &                    To confirm the appointment time and show that he will bring his husband. \\
        & understanding &     19 &              To confirm her understanding of desert classification and provide a visual aid \\
\midrule
introduce & dog &     24 &                                     To introduce her dog to Rance and show how it helps her \\
        & pet &     23 &                                                 To introduce his pet and show how it looks. \\
        & topic &     19 &                     To introduce the topic of baseball and initiate a conversation about it \\
        & cat &     18 &                                           To introduce her cat named after a Cars character \\
\midrule
clarify & difference &     31 &                         To clarify the difference between divorce and annulment for Maxwell \\
        & confusion &     16 &          To clarify Ryley's confusion about Osiel's profession and provide a visual example \\
        & concept &     13 &                                       To clarify the concept of nearsightedness for Conrad. \\
        & misconception &     13 &           To clarify the misconception about black roses and show the actual dark red rose. \\
\midrule
celebrate & achievement &     47 &                           To celebrate her achievement and share her excitement with Shanya \\
        & promotion &      9 &                            To celebrate Britney's promotion and share the news with others. \\
        & birthday &      8 &                                To celebrate Rupert's birthday and make the moment memorable \\
        & accomplishment &      7 &                       To celebrate the accomplishment and share the excitement with Ayelet. \\
\midrule
suggest & activity &     15 &                     To suggest an alternative activity for her kids instead of watching TV. \\
        & place &     14 &                                    To suggest a place to eat and provide a visual reference \\
        & restaurant &     12 &                         To suggest a specific restaurant or location for their next hangout \\
        & solution &      9 &                      To suggest a solution to make up for the lie and mend the relationship \\
\midrule
explain & concept &     24 &                         To explain the concept of two hand touch and flag football visually \\
        & process &     10 &                To explain the process of setting the minimum wage and the parties involved. \\
        & reason &      8 &                     To explain the reason for the stain and show their efforts to remove it \\
\bottomrule
\end{tabular}
\end{adjustbox}
\caption{The top 20 most common root verbs and their up to 4 direct noun objects in the generated rationale. Only pairs with a count of 5 or more are included.}
\label{supp_tab:rationale_distribution}
\end{table*}

\end{document}